# Learning Rhetorical Structure Theory-based descriptions of observed behaviour

Luís Botelho[†1]   Luís Nunes[†]   Ricardo Ribeiro[‡]   Rui J. Lopes[†]

Luis.Botelho@iscte.pt   Luis.Nunes@iscte.pt   Ricardo.Ribeiro@iscte.pt   Rui.Lopes@iscte.pt

† Instituto de Telecomunicações (IT-IUL), Departamento de Ciências e Tecnologias da Informação do Instituto Universitário de Lisboa (ISCTE-IUL)

Av. das Forças Armadas 1649-026, Lisboa, Portugal

‡ Instituto de Engenharia de Sistemas e Computadores (INESC-ID), Departamento de Ciências e Tecnologias da Informação do Instituto Universitário de Lisboa (ISCTE-IUL)

Av. das Forças Armadas 1649-026, Lisboa, Portugal

## Abstract

In a previous paper, we have proposed a set of concepts, axiom schemata and algorithms that can be used by agents to learn to describe their behaviour, goals, capabilities, and environment. The current paper proposes a new set of concepts, axiom schemata and algorithms that allow the agent to learn new descriptions of an observed behaviour (e.g., perplexing actions), of its actor (e.g., undesired propositions or actions), and of its environment (e.g., incompatible propositions). Each learned description (e.g., a certain action prevents another action from being performed in the future) is represented by a relationship between entities (either propositions or actions) and is learned by the agent, just by observation, using domain-independent axiom schemata and/or learning algorithms. The relations used by agents to represent the descriptions they learn were inspired on the Theory of Rhetorical Structure (RST). The main contribution of the paper is the relation family *Although*, inspired on the RST relation *Concession*. The accurate definition of the relations of the family *Although* involves a set of deontic concepts whose definition and corresponding algorithms are presented. The relations of the family *Although*, once extracted from the agent's observations, express surprise at the observed behaviour and, in certain circumstances, present a justification for it.

The paper shows results of the presented proposals in a demonstration scenario, using implemented software.

## 1 Introduction

The capability of understanding or explaining behaviour and its actors is an important feature of computer systems in a large variety of circumstances. In general, explainability, i.e., the system's ability to explain its behaviour, has been one of the most demanded and sought for properties of intelligent systems ([van der Waa et al 2021][Miller 2019][Samek et al 2019]).

In interactive systems, mostly in the contexts of natural language processing and interactive games, an agent may better adapt to its users if it understands their behaviour, mainly their beliefs, motives and preferences (e.g., [Pinhanez et al 2021][Ramirez and Bulitko 2014]).

---

[1] Corresponding Author

Several researchers (e.g., [Köchling and Wehner 2020][Barocas et al 2017]) have reported unacceptable strong biases, which may were not detected by their users, especially in opaque machine learning systems, that would have been noticed if the systems would explain their behaviour. For example, according to [Suresh and Guttag 2021], Northpointe's COMPAS, a model that predicts the likelihood that a defendant will re-offend, exhibited a significantly higher false positive rate for black defendants versus white defendants. The bias would be immediately detected if COMPAS would explain that a defendant had high risk of re-offending because they had a darker skin tone.

The purpose of our research is to improve the degree to which software agents understand observed behaviour (either their own or that of some other agent), its actor and environment. To achieve that, we defined a set of concepts through logical axiom schemata and implemented a set of domain independent algorithms that allow an agent, with no prior explicit understanding of what happens, to acquire such an understanding, just by observation.

In [Botelho et al 2019], we show the way an agent with no prior explicit knowledge about its goals, capabilities, the environment, and its interaction with the world may acquire explicit knowledge of all these concepts. However, the discovered concepts may be extended to enable even richer descriptions. In the present article, we extend the previously proposed set of concepts the agent may discover, and we present concrete results achieved by implemented software in a demonstration scenario.

We want to stress an important difference relative to [Botelho et al 2019]. In the previous article, the main research goal was to empower an agent with the means to better understand itself and the world. Here, we have adopted a more general perspective where an agent observes some behaviour (its own or that of some other agent such as the user), its actor, and the environment in which it unfolds. The set of concepts, algorithms and axiom schemata used by the agent in [Botelho et al 2019] may readily be used by an observing agent to describe observed behaviours, their actors and environment. This article presents an extension of those concepts, algorithms and axiom schemata.

In addition to the mentioned change of perspective, the major contributions of the current extension of our previous work comprise *(i)* the basis for a qualitative scale of preferences towards propositions or actions; *(ii)* two types of incompatibility between propositions and/or actions; *(iii)* perplexity; *(iv)* relation definition; and *(v)* the use of background knowledge for inferring new knowledge. All these contributions mean that an agent learns to better understand the observed behaviour, its actor and environment.

*Qualitative scale of preferences*. Just by observing some behaviour, the observing agent learns some of the attitudes of the actor of the observed behaviour towards its actions and world propositions: desired actions and propositions, undesired actions and propositions, and neutral actions and propositions. Given that a desired action is better than a neutral action, which in turn is better than an undesired action (and the same with respect to propositions), this forms the basis for a qualitative scale of the preferences of the actor of the observed behaviour (e.g., the user preferences).

Note that it is impossible to ensure that the attitudes the observing agent learns of the actor of the observed behaviour actually correspond to first-person attitudes of the observed actor. The observing agent just learns that, according to its own view, the observed behaviour is consistent with its actor having the learned attitudes.

*Incompatibility between propositions and/or actions*. Just by observing some actor's behaviour, the agent learns pairs of incompatible propositions about the world (they cannot be simultaneously true) and it also learns pairs of propositions where one of them being true in a certain state of the world prevents the other from ever becoming true in the future. The incompatibility between world propositions is used, for example, in classical planning algorithms [Ghallab, Nau and Traverso 2004], for instance for detecting incompatible sets of goals but, in some cases, it is expected that the agent developer manually provides such knowledge. It is easy to accept that the knowledge that a proposition prevents another from ever becoming true in the future could also be used in planning the agent's action. As it will be seen, incompatibility and prevention apply to both two propositions, two actions, or a proposition and an action.

*Perplexity*[2]. Maybe, the most interesting innovation of this paper, certainly the most sophisticated, is learning to detect perplexing actions just by the observation of behaviour. In the present article, perplexity is formalized in deontic terms relying on the rigorous definition of ideality principles. Perplexity, surprise and awe are epistemic emotions in the sense that they motivate the agent to seek the new knowledge required to understand the cause of the emotion (e.g., [Deckert and Koenig 2017] [de Cruz 2021] [Vogl et al 2019][McPhetres 2019]). These emotions may thus be used in artificial agents to trigger knowledge seeking or otherwise exploratory behaviours.

Another important note must be presented regarding perplexity. Since it is the observing agent that is actively acquiring information about the observed behaviour and processing it, it is the observing agent that becomes perplexed with particular actions of the actor of the observed behaviour. However, the actor of the observed behaviour might have exactly the same reasons to be perplexed about its own actions if it were aware of what it is doing. Thus, this contribution of our work may apply both to an agent that is observing some behaviour (possibly its own) or to the actor of the behaviour it is observing.

*Relation definition*. Just by observing a behaviour, either its own or another one's, the agent learns to define a proposition as a conjunction of other propositions. This knowledge constitutes a deeper understanding of the observed behaviour.

The note we have just presented about perplexity also applies to proposition definition. The observing agent learns the definition of a certain proposition. However, if the actor of the observed behaviour is also aware of what happens in its own world, it may also have learned the same definition.

*Background knowledge*. When an agent learns a certain proposition (possibly a relation between actions and/or propositions), it may have needed its own background knowledge in addition to its observations. If this is the case, the agent creates a fact of the relation *background*, stating the previous knowledge that it used to infer the learned proposition. Albeit none of the descriptions that the agent learns from the observed behaviour required prior knowledge, we provide a formal inference rule that the agent could have used to derive propositions of the *background* relation if it had needed prior knowledge. Being aware that, to infer a certain knowledge, the agent requires prior knowledge constitutes a profound understanding of its relation with its observations.

All the briefly presented contributions, which allow an agent to learn descriptions of observed behaviours, their actors and environments, were inspired from an analysis of the Rhetoric Structure Theory (RST) by Mann and Thompson (1987; 1988).

William Mann and Sandra Thompson [Mann and Thompson 1987] [Mann and Thompson 1988] proposed the Rhetorical Structure Theory (RST), which includes a set of relational concepts that can be used to describe the rhetorical structure of discourse. RST has been used in computational linguistics, for both language understanding [Marcu 2000] [Uzêda, Pardo and Nunes 2010] and discourse generation [Mann 1984] [Kosseim and Lapalme 2000]. Given the proven power of the proposed RST concepts to describe discourse, we have decided to evaluate the possibility of using them to describe behaviour, its actor and relevant properties of its environment. Next section presents our analysis of the Rhetorical Structure Theory for the purpose of capturing the agent's understanding of observed behaviour. We conclude that several concepts proposed in [Botelho et al 2019] may be equated with some of those of RST. Additionally, we have identified some other RST concepts that are proposed in the current paper to extend the degree to which an agent understands observed behaviour.

---

[2] Not to be confused with perplexity as it is used in language models (i.e., analytic measure of uncertainty derived from the branching factor) and in machine learning (e.g., prediction error, model uncertainty)

Section 3 describes our current proposal and presents results achieved by implemented software in a demonstration scenario. All new concepts are rigorously defined. Deduction axiom schemata and domain independent algorithms are described.

Section 4 relates our work with relevant literature. We conclude that a significant subset of the concepts we propose (in [Botelho et al 2019] and in the current article) is also used in related research.

Finally, section 5 presents conclusions and directions for future research.

# 2 Discussion of RST

This section discusses the usefulness of the relations proposed in the Rhetorical Structure Theory (RST) for our research objectives, namely for improving the degree to which an agent understands observed behaviours, their actors and environments. We conclude that some of the RST relations were already addressed in a previous article presenting our research [Botelho et al 2019]; and that some other RST relations, not previously addressed in our research, will allow agents to develop a richer understanding of what they observe.

## 2.1 RST Relations

RST relations are defined to hold between non-overlapping text spans, called the *Nucleus* and the *Satellite*, but there are a few exceptions. For example, both arguments of the relation *Contrast* are nucleus text spans, and the relation *Sequence* applies to more than two text spans.

The purpose and ingredients of RST are different in many respects to the purpose and fundamental ideas of agent understanding. Of those differences, the following ones deserve being emphasized:

- The purpose of RST is text analysis and generation, taking into account communicative intentions; our research has the purpose of improving the understanding the agent develops of an observed behaviour, its actor and environment;
- The main intentions represented in the relations of the RST theory are communicative intentions of the speaker / writer; the intentions of the actor of the observed behaviour (if any) are those motivating their behaviour;
- The distinction between nucleus and satellite is an important aspect of RST; this distinction is not important for our research;
- In RST, sometimes the same text span may arbitrarily represent actions / activities or situations; the concepts expressing the agent understanding predicate actions and/or propositions;
- Most RST concepts are represented by binary relations; the agent understanding is represented by relations and functions of differing arities.

Given the differences between the purpose and objects of the Rhetorical Structure Theory and describing the agent's understanding of an observed behaviour, its actor and environment, we stress that we will consider the RST relations only as a source of inspiration to extend the set of relations we previously proposed in [Botelho and al 2019].

RST proposes the set of relations presented in Table 1. Some of them have only presentational purpose. Others address the subject matter of the discourse.

**Table 1 – RST relations (adapted from [Mann and Thompson 1988])**

| | |
|---|---|
| Circumstance (subject matter) | Antithesis and Concession (presentational) |
| Solutionhood (subject matter) | • Antithesis |
| Elaboration (subject matter) | • Concession |
| Background (presentational) | Condition and Otherwise (subject matter) |
| Enablement and Motivation (presentational) | • Condition |
| • Enablement | • Otherwise |
| • Motivation | Interpretation and Evaluation (subject matter) |
| Evidence and Justify (presentational) | • Interpretation |
| • Evidence | • Evaluation |
| • Justify | Restatement and Summary (subject matter) |
| Relations of cause (subject matter) | • Restatement |
| • Volitional cause | • Summary |
| • Non-volitional cause | Other relations (subject matter) |
| • Volitional result | • Sequence |
| • Non-volitional result | • Contrast |
| • Purpose | |

Albeit the relations about the subject matter may seem more likely to be useful for the purpose of our research, we will also analyse presentational relations.

The following sections describe the meaning of all RST relations, organizing them according to their possible usefulness for the purpose of representing the understanding of the agent about an observed behaviour, its actor and its environment.

The examples of all RST relations are taken from [Mann and Thompson 1988], except when otherwise specified.

### 2.1.1  Alternative and sequence

The notions of action sequence and of alternative ways to handle a given problem usefully describe agent behaviour. This section shows that RST relations *Sequence* and *Contrast* capture those or very similar concepts.

RST *Contrast* presents two contrasting actions or activities, or situations that may result of those actions or activities. The two text spans of a *Contrast* relationship do not present incompatible situations. The relationship does also not present the two situations as seemingly incompatible. However, the two situations contrast in some respect. In a sense, as the following example shows, *Contrast* can be used to express alternative ways of solving some problem:

> *Animals heal, but trees compartmentalize. They endure a lifetime of injury and infection by setting boundaries that resist the spread of the invading micro-organisms.*

It can be relatively easy to understand that two approaches to solve a similar problem are different, but it would be much more difficult to understand that they somehow contrast. We recognize the importance of proposing relations with those two connotations (i.e., alternative and contrasting). However, we are not ready yet for providing an algorithm capable of distinguishing contrast from mere difference.

RST *Sequence* represents a sequence of the actions or situations presented in the several related text spans. Our approach to describe the observed behaviour captures its sequence of states by the predicate *NextState/3*, which relates a given state of the considered behaviour, the action executed in that state, and the next state, resulting of executing that action.

## 2.1.2 Action effects and relevant achievements

Several RST relations (i.e., *Solutionhood*, *Purpose*, *Enablement*, *Motivation*, *Volitional cause*, *Volitional result*) capture a similar notion that one or several actions may be executed to achieve a given state of affairs.

One of the text spans of the *Solutionhood* relation presents a problem. The other text span represents an action that solves the presented problem. In the following example, "to redistribute the fillers" is a solution to the "insulation tendency to slip towards the bottom".

> *One difficulty with sleeping bags in which down and feather fillers are used as insulation. This insulation has a tendency to slip towards the bottom. You can redistribute the filler.*

Clearly, the RST *Solutionhood* represents the concept that an action was or may be executed to achieve a given state of affairs, such as a goal.

RST *Purpose* also captures a similar relation between two text spans. However, RST *Purpose* is more accurately interpreted as a relation among behaviours than as a goal / action relation. In the following example, "becoming as tall as possible" is the way to enable the realization of photosynthesis:

> *Presumably, there is a competition among trees in certain forest environments to become as tall as possible so as to catch much of the sun as possible for the photosynthesis*

One of the text spans of the RST relation *Volitional cause* presents a volitional action or a situation that could have resulted of a volitional action. The other text span represents a situation that could have caused the agent of the volitional action to perform it. In the following *volitional cause* example, writing being impossible caused the decision to have the typewriter serviced:

> *Writing has almost become impossible so we had the typewriter serviced and I may learn to type decently after all these years.*

This example may be framed as the rule *If you want to learn to type decently (and the typewriter is not working) then you must have the typewriter serviced.*

RST *Volitional-result* is the exact symmetric of RST *Volitional-cause* in the sense that, in *Volitional-result*, it is the nucleus text span that causes the satellite text span, while in *Volitional-cause*, it is the satellite that causes the nucleus.

The RST relations *Enablement* and *Motivation* also suggest the concept that an action can be executed to achieve a given state of affairs, but the purpose is different since both are concerned with enabling or motivating the reader to perform a certain action.

One of the text spans of the RST *Enablement* represents an unrealized action (to be performed by the reader). The other text span presents information meant to motivate the reader to do the action. In the following RST *Enablement* example, the information regarding training on jobs is presented to motivate the reader to ask for a catalogue and order form:

> *Training on jobs. A series of informative, inexpensive pamphlets and books on worker health discusses such topics as filing a compensation claim, ionizing radiation, asbestos, and several occupation diseases. For a catalog and order form write to WIOES, 2520 Milvia St., Berkeley, CA 95704.*

This example can be cast as the rule *If you want to enrol the employees in training programs (which are important to learn about filing a compensation claim, ionizing radiation, asbestos, and several occupational diseases) then get a catalogue and order form from WIOES, 2520 Milvia St., Berkeley, CA 95704.*

RST *Motivation* captures a similar relation between a text span that is meant to motivate the reader to perform the action presented in the other text span.

Two related, but different, concepts are suggested by the described RST relations: the general effects of an action, independent of specific executions, and the relevant effects of an executed action, in the scope of a specific behaviour. These two concepts are handled by the functions *PosEffects/1* and *NegEffects/1* and by the relation *Achieved/3* of our original proposal [Botelho et al 2019]. *PosEffects/1* and *NegEffects/1* represent the general (positive and negative) effects of an action, independent of any specific execution. *Achieved/3* represents the relevant effects of an action executed in the scope of a particular behaviour. We use the expression "relevant effects" to refer to the propositions that were not true in the state in which the action was executed and became true after the action was executed. They are relevant because they are either goals or enable the execution of an action that was actually executed in the considered behaviour.

### 2.1.3 Contribution and precondition

While section 2.1.2 is focused on action effects, this section addresses different action preconditions, which are captured by several RST relations (i.e., Condition, Non-volitional cause, Non-volitional result).

The situation described in one of the text spans of the RST relation *Condition* can only be achieved if the situation described in the other text span is achieved before. In the following example (adapted from [Mann and Thompson 1988], but not exactly the same), the new spouse or children receiving benefits can only happen if the employee has completed new beneficiary designation forms:

> *Employees are urged to complete new beneficiary designation forms for retirement or life insurance benefits whenever there is a change in marital or family status. This is the only way new spouses or children may receive benefits.*

The accurate interpretation of the RST *Condition* requires a more detailed analysis. *Condition* relates a situation, that is, a set of propositions ($PropSet_2$) that can only be achieved if another set of propositions ($PropSet_1$) is achieved before. $PropSet_1$ is the condition for $PropSet_2$. Implicitly, the relation assumes the existence of an (unmentioned) action (say $A_{1.2}$) that would result in the achievement of the second set of propositions. That unmentioned action $A_{1.2}$ can only be performed if the set of propositions $PropSet_1$ is the case. This analysis shows that RST *Condition* represents a relation between an unmentioned action and the set of propositions that must be true so that the action can be executed.

The expression *non-volitional* of the two RST relations *Non-volitional cause* and *Non-volitional result* is used to express the notion of a situation that enables an action to be executed. That is, the referred situation is not the motivation for executing the action. Instead, the referred situation is the opportunity for its execution.

According to its definition, one of the text spans of the RST relation *Non volitional cause* presents a non-volitional action or a situation that could have resulted of a non-volitional action. The other text span represents a situation that could have caused the agent of the non-volitional action to perform it. In the following *Non-volitional cause* example, being able of mining more than necessary causes the possibility of exporting:

> *We have been able to mine our own iron ore, coal, manganese, dolomite, all the materials we need to make our own steel. And because we can mine more than we need, we've had plenty manganese and iron ore for export.*

The action of exporting is non-volitional because it was not intended *a priori*; it became an option once it was realized that more than enough manganese and iron ore was mined.

Being able of mining more than necessary does not cause the action. The unmentioned goal of increasing profit causes the action of exporting. Being able of mining more than necessary is the condition that enables exporting. However, if the company did not want to make more money, they would not have bothered exporting.

RST *Non-volitional-result* is the exact symmetric of RST *Non-volitional-cause* in the sense that, in *Non-volitional-result*, it is the nucleus text span that enables / contributes to the satellite text span, while in *Non-volitional-cause*, it is the satellite that enables / contributes to the nucleus.

The discussed RST relations suggest two different, albeit related concepts: the general preconditions of an action, independent of specific executions, and the contribution of a certain proposition to an executed action, in the scope of a specific behaviour. These two concepts are handled by the function *Precond/1* and the relation *Contributed/3* of our original proposal [Botelho et al 2019]. *Precond/1* represents the general preconditions of an action, independent of any specific execution. *Contributed/3* represents the relevant contribution of an observed proposition to an action executed in the scope of a particular behaviour.

### 2.1.4 Incompatibility and prevention

The RST relations *Antithesis* and *Otherwise* suggest two important concepts: incompatibility and prevention. If an agent learns that two propositions are incompatible or that one of them prevents the other from holding in the future, it gains a richer understanding of the environment. If the observing agent is the actor of the observed behaviour, it may use the learned knowledge to make better decisions on how to act in that environment.

In the RST relation *Antithesis*, one cannot have positive regards for both the nucleus and the satellite. In the following example of the RST *Antithesis*, the lack of jobs is presented as an antithesis of laziness.

> *The tragic and too-common tableaux of hundreds or even thousands of people snake-lining up for any task with a paycheck illustrates a lack of jobs, not laziness.*

We will propose the *Incompatible/2* relation to express the incompatibility of two domain entities.

The RST relation *Otherwise* means that if the situation described in one of the text spans is true, the situation described in the other text span cannot be achieved. *Otherwise* captures a prevention relationship between propositions. We will add the relation *Prevents/2* to capture the idea that the occurrence of one of the domain entities will prevent the future occurrence of another domain entity.

### 2.1.5 Definition / Restatement

RST relations *Summary* and *Restatement* represent different ways of expressing the same idea or situation. RST relation *Restatement* holds of two text spans if one of them is a restatement of the other. However, the two text spans must be of comparable bulk. In the following example of the RST *Restatement*, the second statement expresses exactly the same idea as the first one:

> *A well-groomed car reflects its owner. The car you drive says a lot about you.*

RST relation *Summary* holds of two text spans when one of them is a shorter restatement of the other. In the following example of the RST *Summary*, improving computer performance is a summary of improving memory and getting a big bonus in computer performance:

> *For top quality performance from your computer, use the flexible disks known for memory excellence. It's a great way to improve your memory and get a big bonus in computer performance.*

The RST relations *Restatement* and *Summary*, exactly as used in RST, don't apply to our research goal. However, two different concepts suggested by these two RST relations are important concepts to express the agent understanding of observed behaviour, its actor and environment. One is the idea of alternative ways to achieve the same goal, as already mentioned with respect to the RST relation *Contrast* (section 2.1.1). The other is the concept of a proposition that is defined in terms of other propositions. In this last case, it might be said that the defined proposition represents a shorter way of expressing the defining set of propositions.

Defining a proposition in terms of other propositions may also capture the hierarchic organization of the agent behaviour. For instance, *"to achieve its goal, the agent ensures that the stack is inside the closet and that the closet door is closed"* is a more abstract (in the sense of less detailed) version of *"to achieve its goal, the agent ensures that block B is on top of block C and block A is on top of block B forming a stack, that the stack is inside the closet, and that the closet door is closed"*.

In this paper, we propose the relation *Defining/2* as a way of expressing that a proposition is defined as the conjunction of other propositions. The paper provides an algorithm used by the agent to acquire definitions of propositions from the observed behaviour, and results of an actual implementation.

### 2.1.6 Concession

The RST relation *Concession* is well understood as a synonym for *Although*, for example

> *Although it is toxic to certain animals, evidence is lacking that it has any serious long-term effects on human beings.*

We will propose two versions of *Although*, based on the RST relation *Concession*. The new proposals are used by the agent to express perplexity with some action. In one of the versions, the agent also expresses the reason why the perplexing action has been executed. Perplexity and the rational for a perplexing action represent a deep understanding of the considered behaviour.

The two versions of the relation *Although* will be one of the most important aspects of this article. Rigorous definitions and algorithms used by the agent to acquire relationships of the family *Although* from the observed behaviour will be provided along with results of an actual implementation.

### 2.1.7 Enabling or improving an interpretation

Three RST relations (i.e., *Background*, *Circumstance* and *Interpretation*) are used when one of the text spans introduces new information that enables or improves the understanding of the other text span.

One of the text spans of the RST relation *Background* provides background information that increases the reader's understanding of the other related text span. In the following example of the RST relation *Background*, the previous Government Code presents background information necessary for a complete understanding of the new bill that protects public employees' personal data:

> *Home addresses and telephone numbers of public employees will be protected from public disclosure under a new bill approved by Gov. George Deukmejian. Assembly Bill 3100 amends the Government Code, which required that the public records of all state and local agencies, containing*

> *home addresses and telephone numbers of staff, be open to public inspection.*

One of the text spans of the RST relation *Circumstance* sets a framework within which the reader is intended to interpret the situation presented in the other related text span. In the following example of the RST *Circumstance*, the text span that presents the Prime Minister volunteering to work at KUSC[3], while attending the Occidental College in 1970, defines the circumstances that enable a better understanding of the statement that the P.M. has been with KUSC longer than any other staff member:

> *P.M. has been with KUSC longer than any other staff member. While attending Occidental College, where he majored in philosophy, he volunteered to work at the station as a classical music announcer. That was in 1970.*

One of the text spans of the RST relation *Interpretation* links the situation presented in the other text span to a framework of ideas not involved in that text span and not concerned with the writer's positive regard. In the following example, the unusualness of the decline mentioned in the second text span is related to the decline of the composite mentioned in the other text span:

> *Steep declines in capital spending commitments and building permits, along with a drop in the money stock pushed the leading composite down for the fifth time in the past 11 months to a level 0.5% below its high in May 1984. Such a decline is highly unusual at this stage in an expansion.*

In the three RST relations, one of the text spans provides additional information that, when related with the information presented in the other text span, allows a better understanding of the other text span. Both relations suggest the usefulness of identifying a bulk of possibly declarative knowledge or theory that would improve or enable the correct understanding of the agent behaviour. In this paper, we propose the *Background* relation as a way of expressing the relationship between *a priori* knowledge of the domain and concepts extracted by the agent from an observed behaviour, its actor and environment. Our research proposal explicitly tries to avoid requiring the agent to use *a priori* knowledge to understand the observed behaviour. However, shall necessity arise, the newly proposed *Background* relation can be used.

### 2.1.8 Evaluation

One of the text spans of RST *Evaluation* relates the situation presented in the other text span to the degree of the writer's positive regard towards it. The intended effect of an evaluation relation is that the reader recognizes that one of the text spans assesses the situation presented in the other text span and recognizes the value it assigns to it. In the following example *of the relation Evaluation*, better performance and reliability is an assessment of features of the referred disk:

> *Features like our uniquely sealed jacket and protective hub ring make our disks last longer. And a soft inner liner cleans the ultra-smooth disk surface while in use. It all adds up to better performance and reliability.*

Evaluating the observed behaviour or alternative courses of action reflects a deep understanding of what is observed. If the observing agent is the actor of the observed behaviour, the capability to evaluate may allow the agent to improve its behaviour.

We propose the new relation *Evaluation* to express the agent evaluation of possible courses of action.

---

[3] KUSC is a classical music radio based on the University of Southern California.

### 2.1.9 Evidence and justification

Two RST relations (*Evidence* and *Justify*) are concerned with presenting information that can be seen as evidence or justification for the presentation of other information.

In RST *Evidence*, one of the text spans is presented to increase the reader's belief in the other text span. In the following example, the second topic is presented to increase the reader's belief in the first topic:

> 1. *The program published for calendar year 1980 really works*
> 2. *In only a few minutes, I entered all the figures from my 1980 tax return and got a result which agreed with my hand calculation to the penny.*

The RST relation *Evidence* can be cast as a condition-conclusion rule. The presented example can be stated as the following *if then* rule:

> *If using a program requires only a short time, and the program provides a comprehensive coverage of the target situation, and the program results are as expected then the program really works (mostly in terms of usability).*

Condition-conclusion rules can actually be interpreted as meaning that if the condition is believed then the conclusion is also believed, that is, the conditions works as evidence for the conclusion.

The RST relation *Justify* has a slightly different purpose. One of the related text spans is used to increase the reader's readiness to accept the writer's right to present the other text span. In the following example of the relation *Justify*, the second and third topics are presented to increase the reader's readiness to accept the writer's right to present the first topic:

> 1. *The next music day is scheduled for July 21 (Saturday), noon-midnight*
> 2. *I'll post more details later,*
> 3. *But this is a good time to reserve the place on your calendar*

RST *Justify*, exactly as it stands, is not useful for our research purposes because the goal of using the relation is to justify the writer's right to present a certain piece of information. A similar albeit different idea would be worth considering, the idea that a given state of affairs (a set of propositions) justifies the performance of an action. A set of propositions is a justification for an action if those propositions are goals the actor of the observed behaviour wants to achieve and the action actually achieves them, or if the propositions achieved by the action are preconditions for future, also justified, actions. This is exactly the intent of our original relation *Achieved/3*, which expresses the relation between an action and its relevant results. A result is relevant if it is a goal of the agent or if it is a precondition of another action executed posteriorly. *Achieved/3* is discussed in section 2.1.2.

### 2.1.10 Elaboration

The RST relation *Elaboration* is used to present more details about a certain situation. One of the text spans of RST *Elaboration* presents more detail to the situation presented in the other text span. In the following example of RST *Elaboration*, the description of the expected attendees and the information regarding the topics of the conference (one of the text spans) provide additional details about the conference presented in the other text span:

> *Sanga-Saby-Kursgard, Sweeden, will be the site of the 1969 International Conference on Computational Linguistics, September 1-4. It is expected that some 250 linguists will attend from Asia, West Europe, East Europe including Russia, and the United States. The conference will be concerned with the application of mathematical and computer techniques to the study of natural languages, the development of computer programs as tools for* *linguistics research, and the application of linguistics to the development of man-machine communication systems.*

The RST relation *Elaboration* suggests two different, although related concepts: *(i)* describing a behaviour in more detail, and *(ii)* defining a proposition as a set of propositions. This second concept, which is more superficially linked to the RST *Elaboration*, has already been considered in section 2.1.5.

The first concept, namely that of describing a certain behaviour in more detail, is the nature of complex behaviours. Complex behaviours have ultimate goals to be achieved. Those goals are achieved through the achievement of other more specific instrumental goals, each of which may also involve other yet more specific instrumental goals, and so forth. Often, each goal (instrumental or not) may be fulfilled through alternative combinations of other goals. This corresponds to the recognition of the hierarchic structure of complex behaviour. The hierarchic nature of complex behaviour is already captured by the relations *Achieved* (section 2.1.2) and *Contributed* (section 2.1.3), originally proposed in [Botelho et al 2019], and *Defining* (section 2.1.5), presented in this article. The possibility of alternative approaches to achieve a given state of affairs was discussed in section 2.1.1.

## 2.2 Extending our original proposal

This section briefly describes the set of concepts that resulted from the analysis of the RST relations presented in section 2.1. Then, we contrast the referred set of relations with the relations we originally proposed in [Botelho et al 2019]. We show that several of the RST-based concepts are captured by our original proposal and we identify those that were not yet covered.

The RST-based concepts resulting from the analysis in section 2.1 are listed in Table 2.

**Table 2 – RST-based concepts**

| Sequence | Effects (Positive and negative effects) |
|---|---|
| Alternative | Precondition |
| Achieved | Incompatible |
| Contributed | Prevents |
| Although | Defining |
| Evaluation | Background |
| Evidence | |

*Sequence*. An observed ordered collection of states forms a sequence in a specific behaviour. The predicate *NextState/3*, directly built from the sensors of the observing agent, captures the notion of state sequence.

*Alternative*. Specific behaviours $B_1$ and $B_2$ are alternative ways of achieving the same results. Our original proposal does not capture this concept. For that, we would have to observe different behaviours to handle the same problem, which we did not. This paper will not address this possibility. We leave it for future research.

The relations *Achieved* and *Contributed* express properties of actions and propositions of specific behaviours. The former is used to express the fact that a certain action, executed in a certain state of a certain behaviour, achieved a certain set of relevant effects. The latter is used to express the fact that a certain proposition, holding in a specific state of a specific behaviour, contributed to enabling the execution of a specific action (because it is one of its preconditions).

The word *Although* may be used to express different classes of relations. In the example provided in the RST paper [Mann and Thompson 1988],

> *Although it is toxic to certain animals, evidence is lacking that it has any serious long-term effects on human beings*

the word is adequately used because of the perplexity that is caused by the fact that the effect of some action is one for certain animals but another one for other animals.

We will restrain the use of the relation *Although* to express the fact that although the specific action executed by the observed actor, in a specific state, seems to deviate it from an apparently more favourable state, in terms of the degree of satisfaction of a given ideality principle, it was nevertheless executed. A second version of the relation will allow the observing agent to describe the justification it has learned for the execution of the apparently perplexing action.

The relation *Evaluation* may be used by the observing agent to assign a certain evaluation to a specific course of action or to a specific situation. In case the observer agent is the actor of the observed behaviour, evaluating alternative courses of action will enable the agent to choose among possible actions to be performed. Although recognizing its importance, we do not explicitly handle the relation *Evaluation*. However, in section 3.1, we discuss and formalize diverse concepts well related with *Evaluation*. The learned predicates *Desired/1*, *Neutral/1* and *Undesired/1* directly evaluate their arguments (propositions or actions). Additionally, we provide the means for the observing agent to infer the degree to which several ideality principles are satisfied in a given state of an observed behaviour. Given that an ideality principle reflects the way the behaviour should unfold, determining the degree to which an ideality principle is satisfied is an implicit evaluation of that behaviour.

The relation *Evidence* may be used to express the fact that a certain specific proposition is evidence that another specific proposition is true. Although recognizing its importance, we do not tackle the relation *Evidence*.

*Action effects*. Actions, if successfully executed, have effects on the world. We use the functions *PosEffects/1* and *NegEffects/1* to represent the positive and the negative effects of an action. The positive effects of an action are those propositions that become true after the action is executed. The negative effects of an action are those propositions that cease to be true after the action is executed.

*Action preconditions*. For an action to be executed, certain conditions must hold. We use the function *Precond/1* to represent the set of propositions that must hold for the action to be executed.

The relation *Incompatible* will be used to represent the incompatibility between two domain entities. Two entities are incompatible if they cannot occur in the same state.

The relation *Prevents* will be used to express the fact that one domain entity (either a proposition or an action), once occurring in a given state of a given behaviour, prevents the future occurrence of another entity in the same behaviour.

The relation *Defining* is used to represent the definition of one proposition as the conjunction of a set of other propositions.

Finally, the relation *Background* can specify the *a priori* background knowledge used by the agent to acquire a certain relationship between world propositions and/or actions.

After having briefly described the RST-based relations, useful for representing the agent learned description of observed behaviour, its actor and environment, we present the relations originally proposed in [Botelho et al 2019] with the same purpose. Several of the RST-based relations represent the same concepts as those we have already proposed. However, there is a significant set that was not covered in our previous research.

Some of the previously defined descriptions are properties of the observed entities of the domain (Table 5, e.g., *Proposition/1*, *Action/1* and *Goal/1*). The other previously proposed descriptions relate entities of the domain (Table 4 and Table 3, e.g., *Achieved/3*, *Contributed/2*, *MustPrecede/2*, *PosEffects/1* and *NegEffects/1*). In this context, domain entities comprise actions and propositions.

The relations we have previously proposed were organized into two groups according to whether they apply only to specific behavioural instances (Table 3, e.g., *Achieved/3* and *Contributed/2*) or to all behaviours of the same class (Table 4, e.g., *MustPrecede/2* and *Precond/1*).

Finally, some descriptions were represented as functions (e.g., *PosEffects/1* and *NegEffects/1*) whereas others were represented as predicates or relations (e.g., *Achieved/3*, *MustPrecede/2*).

We start with the relations that may be used to describe specific behaviours. Then we move to those that apply to the whole class of behaviours. Finally, we present those relations that predicate individual entities of the domain.

**Table 3 – Descriptions that relate entities. Applicable to specific behaviours**

| NextState(State$_1$, Action, State$_2$) | The predicate *NextState/3* relates a state identifier (*State$_1$*) with the action that was executed in that state (*Action*), and the resulting state (*State$_2$*). *NextSate/3* is defined only for observed actions of specific behaviours. |
|---|---|
| Achieved(State, Act, PropsSet) | The propositions in the set *PropsSet* constitute the relevant effects of the action *Act*, executed in the state *State*. *Relevant* in the sense that each of them is either a goal of the actor of the observed behaviour or one of the preconditions of an action that actually used it in a future sate of the same behaviour. |
| Contributed(State, Prop, Act) | Proposition *Prop*, true in the state *State*, contributed to the execution of the action *Act* in the state *State*, because it is one of *Act*'s preconditions. |

Table 3 shows that RST-based relations *NextState*, *Achieved* and *Contributed*, which were already included in our previous work [Botelho et al 2019].

Several concepts relating domain entities are applicable to all instances of the same class of behaviours. For instance, the set of effects of an action describes the action, not a specific execution of that action in a specific behaviour.

**Table 4 – Descriptions that relate entities. Applicable to all behaviours**

| MustPrecede(Prop$_1$, Prop$_2$) | In all instances of the same class of behaviours in which *Prop$_1$* and *Prop$_2$* have occurred, *Prop$_1$* and *Prop$_2$* occur both in the initial state or *Prop$_1$* must occur before *Prop$_2$*. <br><br> The precedence relation does not arise of *Prop$_1$* being a precondition of any of the actions that lead from the state in which *Prop$_1$* is true to the state in which *Prop$_2$* is true. |
|---|---|
| Precond(Act) | *Precond(Act)* is the set of preconditions of the action *Act*, in the sense that *Act* can only be executed in a state in which all of its preconditions are true. |
| PosEffects(Act) | *PosEffects(Act)* is the set of positive effects of *Act*, that is, the set of propositions that become true in the state immediately after *Act* is executed. |
| NegEffects(Act) | *NegEffects(Act)* is the set of negative effects of *Act*, that is, the set of propositions that cease to be true in the state immediately after *Act* is executed. |
| ValidityCondition(Act) | *ValidityCondition(Act)* is the validity condition of the action *Act*. The validity condition of an action specifies the set of valid instantiations that can be applied to the action variables. |

Table 4 presents the RST-based relations *Precond*, *PosEffects*, and *NegEffects*, which were already part of our previous work [Botelho et al 2019].

In addition to the relations between domain entities, we also use a set of unary relations that represent properties of domain entities.

**Table 5 – Entity properties**

| Proposition(Entity) | *Entity* is a proposition |
|---|---|
| StaticProposition(Entity) | *Entity* is a static proposition (true in all states of all behavioural instances of the same class). |
| FluentProposition(Entity) | *Entity* is a fluent proposition (true only in some states) |
| Action(Entity) | *Entity* is an action. |
| Goal(PropsSet) | The goal of the actor of the observed behaviour is the conjunction of all propositions of the set *PropsSet*. That is, each behavioural instance of the same behavioural class is directed to achieve a state in which all propositions in *PropsSet* are true. |
| Desired(Prop) | *Prop* is a desired entity. A proposition is desired if it is one of the agent's goals. |
| Mandatory(Prop) | *Prop* is a mandatory proposition in the sense that, although it is not a goal, there has to be at least one state in the observed behaviour, in which the proposition is true, otherwise the actor of the observed behaviour won't achieve their goals. The obligation results of the interaction of the task structure, the actor's goals and their capabilities. |

Since there are no unary RST relations, none of them is similar to the previously presented relations expressing properties of domain entities.

Some RST relations apply indistinctly to both propositions and actions. However, the rigorous meaning of a relation depends on the nature of the entity (proposition / action) to which it applies. One way of dealing with this problem would be to define different versions of the same relation depending on the nature of its arguments. The other way is to have only one version of each relation but providing a way of expressing the nature of its arguments. Our relations *Proposition/1*, *StaticProposition/1*, *FluentProposition/1* and *Action/1* are used to distinguish actions from two different kinds of propositions.

From the comparison of our original proposal with the RST-based relations identified in this paper we recognize the desirability to extend our original proposal with the following RST-based relations: *Alternative*, *Although*, *Evaluation*, *Evidence*, *Incompatible*, *Prevents, Defining,* and *Background*. Additionally, we decided to add the unary relations *Undesired/1* and *Neutral/1*, which represent properties of domain entities, because they happen to be important for the rigorous definition of the new added relations (section 3.1). Undesired(Entity) is true if the specified entity is not desired by the agent. Neutral(Entity) means that the specified entity is neither desired nor undesired by the agent.

Of the whole set of newly proposed relations, this paper focuses exclusively on *Undesired, Neutral*, *Incompatible*, *Prevents*, *Defining, Although* and *Background*, and on an extension of the relation *Desired.* In our previous work, the relation *Desired/1* applied only to propositions. Here, we extend its definition to apply to actions as well. The others are left for future work. Section 3 describes our current proposals and achieved results. The next section compares our choices and algorithms with related research.

# 3 Research approach and results

The research presented in this section consists of a set of domain independent algorithms and sometimes axiom schemata that may be used by an agent to learn (sometimes, to derive) descriptions of the observed behaviour, its actor and environment. The descriptions to be acquired by the agent through observation are those identified in section 2.2.

This research relies on some assumptions. First, the learning agent is capable of observing the behaviours under consideration. Each behaviour is associated with a sequence of states connected by the actions performed by its actor on the environment. Each state is a set of propositions about the environment. The first state of each observed behaviour is the initial state. When an action is performed in a certain state, by the actor of the observed behaviour, the world changes from the state in which the action was executed to the next state. Each behaviour ends when the goal of its actor is achieved. This means we also assume that all behaviours are successful, in the sense that the goals of their actors are achieved in the final states of their behaviours. We are aware that this is a limitation. However, this assumption is perfectly realistic in a large class of tasks that are always successfully performed.

The definitions and explanations that follow often rely on the notions of behavioural class, behavioural instance, state, proposition and action. A behavioural instance is a specific timely localized behaviour with a specific set of initial conditions. A class of behaviours is a set of behavioural instances with the same purpose (the goal of the behaviour's actor), although possibly with different initial conditions or different ways of achieving that purpose. Any behavioural instance is a sequence of actions performed by an actor (either the same agent observing and learning from the behaviour or another agent including the user), causing its environment to evolve through a sequence of states.

For convenience, states are indexed by non-negative integers, reflecting the order by which they occur in their specific behavioural instance.

## 3.1 Rigorous definition of the new relations

Following, we present the rigorous definitions, and often deduction axiom schemata, for all the new relations presented in the article. We also provide brief explanations of the algorithms used by the agent to learn descriptions of the observed behaviours. Since the relation *Desired/1* was originally defined for propositions only [Botelho et al 2019], we provide its new definition valid also for actions.

| Desired(Entity) | *Entity* is a desired entity. A proposition is desired if it is one of the goals of the actor of the observed behaviour. An action is desired if its positive effects include desired propositions but no undesired ones, and its negative effects do not include desired propositions. |
|---|---|
| Undesired(Entity) | *Entity* is an undesired entity. A proposition is undesired if it prevents the occurrence of a desired entity. An action is undesired if its positive effects include undesired propositions. |
| Neutral(Entity) | *Entity* is a neutral entity if it is not a desired entity or an undesired entity. |
| Incompatible(Entity$_1$,Entity$_2$) | Entities *Entity$_1$* and *Entity$_2$* are incompatible in the same state, in the sense that they cannot co-occur simultaneously. |
| Prevents(Entity$_1$, Entity$_2$) | The occurrence of the entity *Entity$_1$* in a given state prevents the occurrence of entity *Entity$_2$* in any future state of the same behaviour. |
| Defining(Prop, PropsSet) | Proposition *Prop* is defined as the conjunction of all propositions in the set *PropsSet*. |
| Although(PSet$_1$, A, S, PSet$_2$) | The observed action (*A*) causes perplexity because, although the state of the world before its execution was more favourable for the actor of the observed behaviour than the resulting state *S*, the actor actually executed it. The propositions in *PSet$_1$* and *PSet$_2$* attest the satisfaction degree of the ideality principle, respectively before and after the action was executed. By convention, the ideality principle is included in *PSet$_1$* and in *PSet$_2$*. |
| Although(PSet$_1$, A, S, PSet$_2$, Rational) | *PSet$_1$*, *A*, *S* and *PSet$_2$* have exactly the same interpretation as in *Although/4*. |
| | In spite of the deviation from the ideality principle, the executed action can be justified by *Rational*. The rational for the action execution is a pair of an ideality principle and one of the optimal (shortest) action sequences that leads the actor of the observed behaviour to a state in which that ideality principle (included in the *Rational*) is satisfied. |
| Background(PropsSet, Relationship) | *PropsSet* is background knowledge used by the observing agent to infer the specified *Relationship*. |
| | Ideally, domain background knowledge will not be necessary to acquire the proposed relationships. However, should that be the case, the *Background/2* relation may be used to specify the background knowledge used to infer the relationship. |

### *Desired propositions and actions: Desired/1*

In [Botelho et al 2019], we have presented the definition of desired propositions. The concept of desired actions was left undefined.

The two following deduction axiom schemata may be used to derive instances of the *Desired/1* relation.

$(Goal(PSet) \land P \in PSet) \Rightarrow Desired(P)$

$( Action(A) \land$
 $\exists p\ (p \in PosEffects(A) \land Desired(p)) \land$
 $\neg \exists q\ (q \in PosEffects(A) \land Undesired(q)) \land$
 $\neg \exists r\ (r \in NegEffects(A) \land Desired(r)) ) \Rightarrow Desired(A)$

In the last axiom schema, PosEffects(Act) is the set of all positive effects of Act, and NegEffects(Act) is the set of its negative effects.

The positive effects of an action are those propositions that become true immediately after the action is executed. The negative effects are those propositions that cease to be true after the execution. The functions *PosEffects/1* and *NegEffects/1* were rigorously defined in [Botelho et al 2019].

The algorithm for determining desired actions works as a filter. It selects the observed actions (computed in [Botelho et al 2019]) whose positive effects include at least a desired proposition but no undesired propositions, and whose negative effects do not include a desired proposition.

### *Undesired propositions and actions: Undesired/1*

If a proposition prevents the future occurrence of a desired entity, then it is undesired. This is captured by the following deduction axiom schema:

$\exists e\ (Desired(e) \land Prevents(P, e)) \Rightarrow Undesired(P)$

In which Prevents(Prop, Entity) means that, if the proposition *Prop* is true in a given state of any behaviour, then the entity *Entity* will not occur in any future state of the same behaviour.

Undesired actions are those that give rise to undesired propositions. The following deduction axiom schema captures this definition.

$[Action(A) \land \exists p\ (p \in PosEffects(A) \land Undesired(p))] \Rightarrow Undesired(A)$

The computation of the set of undesired entities follows the straightforward application of the two axiom schemata, which require the relations *Prevents/1* and *Desired/1* and the function *PosEffects/1*.

The algorithm that computes the set of all desired propositions builds the set of propositions *P* such that Prevents(P, E) and *E* is a desired entity.

The algorithm that computes the set of all undesired actions builds the set of actions whose positive effects include at least one undesired proposition.

### *Neutral propositions and actions: Neutral/1*

Neutral entities are those that are neither desired nor undesired. The following two deduction axiom schemata capture the concept.

$(FluentProposition(P) \land \neg Desired(P) \land \neg Undesired(P)) \Rightarrow Neutral(P)$

$(Action(A) \land \neg Desired(A) \land \neg Undesired(A)) \Rightarrow Neutral(A)$

The algorithms for computing the set of neutral entities straightforwardly apply the two axioms by means of set difference operations.

### *Incompatible(Entity1, Entity2)*

Incompatible(Entity$_1$, Entity$_2$) means that entities *Entity$_1$* and *Entity$_2$* cannot occur in the same state of the same behaviour. If *Entity$_1$* and *Entity$_2$* are propositions, they cannot hold simultaneously. If both entities are actions, they cannot be executed in the same state. If one of them is an action and the other is a proposition, then the action cannot be executed in a state in which the proposition holds.

Given that static propositions are not incompatible with any other entity (given that they are true in all states of all behaviours), whenever a proposition is mentioned, it will be a fluent proposition.

Given that each of the two entities *Entity$_1$* and *Entity$_2$* may be a proposition or an action, we present more detailed definitions for the cases of incompatible propositions and for the incompatibility of a proposition and an action. The incompatibility of two actions will not be considered because we assume the actor of the observed behaviour can only execute an action at a time.

In any case, the *Incompatible/2* relation is symmetric.

### *Incompatible propositions*

1. Two propositions are incompatible if one of them is the negation of the other.
2. Two propositions are incompatible if, although none of them is the negation of the other, it is possible to infer the negation of one of them *(i)* from the other and possibly *(ii)* from background or learned domain knowledge.

The goal of our research relative to determining incompatible propositions is that the agent learns, through observation, all pairs of propositions *P$_1$* and *P$_2$* such *P$_1$* and *P$_2$* cannot occur together. It would be possible to add the propositions that result of clause 1 to those learned pairs of propositions, but we have decided not to include negated propositions in the states of the observed behaviour.

### *Incompatible proposition and action*

3. A proposition is incompatible with an action if at least one of the action preconditions is incompatible with the proposition.

Given that the incompatibility of a proposition and an action can be reduced to the incompatibility between propositions, the goal of our research will be the learning of all pairs of incompatible propositions. The incompatibility between an action and a proposition will be derived by the application of the following deduction axiom schema:

[ Action(A) ∧ FluentProposition(P) ∧
 ∃q (q∈Precond(A) ∧ Incompatible(q, P)) ] ⇒ Incompatible(P, A)

in which *Precond/1* is a function that returns the set of preconditions of an action.

The algorithm for the computation of the set of all pairs of incompatible propositions, builds the set of all pairs of fluent propositions <P$_1$, P$_2$>, such that P$_2$ never occurs in a state in which P$_1$ occurs.

The algorithm for computing the set of all incompatible pairs of an action and a proposition consists of the straightforward application of the presented deduction axiom schema, which requires the set of all actions, the set of all fluent propositions, the set of pairs of incompatible propositions, and the relation between actions and their preconditions (represented by function *Precond/1*).

The relations *Action/1* and *FluentProposition/1*, and the function *Precond/1* were rigorously defined in [Botelho et al 2019].

### *Prevents(Entity1, Entity2)*

If entity $Entity_1$ is observed in state $S_1$, then entity $Entity_2$ won't occur in any state $S_2$ after $S_1$. The case in which two entities cannot occur in the same state is represented by the incompatibility relation.

Given that static propositions cannot prevent the future occurrence of another entity, unless the other entity is impossible (which is never the case because we are dealing with observed entities), whenever a proposition is mentioned with respect to the relation *Prevents/2*, we will be talking about a fluent proposition.

If the two arguments of the *Prevents/2* relation are propositions, Prevents($P_1$, $P_2$) means that, if $P_1$ holds in a specific state, then $P_2$ won't hold in any future state of the same behaviour.

If the two entities are actions, then Prevents($A_1$, $A_2$) means that, if action $A_1$ is executed in state $S_1$, then action $A_2$ cannot be executed in a state $S_2$ of the same behaviour occurring after $S_1$.

Being *Prop* a proposition and *Act* an action,

Prevents(Prop, Act) means that, if *Prop* is true in state $S_1$, it won't be possible to execute *Act* in any state $S_2$ of the same behaviour occurring after $S_1$; and

Prevents(Act, Prop) means that, if *Act* is executed in state $S_1$, then *Prop* won't be true in any state $S_2$ of the same behaviour occurring after $S_1$.

*Prevents/2* is neither symmetric nor transitive.

### *Proposition $P_1$ prevents proposition $P_2$*

1. Proposition $P_1$ prevents proposition $P_2$ if it is known, using background domain knowledge or if it is learned by the agent from the observed behaviour

Given that we are not interested in using background domain knowledge, the goal of our research is to learn all pairs of fluent propositions <$P_1$, $P_2$> such that the occurrence $P_1$ prevents $P_2$ from occurring afterwards, in the same behaviour.

### *Action $Act_1$ prevents action $Act_2$*

2. Action $Act_1$ prevents action $Act_2$ if it is known, using background domain knowledge, or if the agent learns it from the observed behaviour.
3. Action $Act_1$ prevents action $Act_2$ if at least one of the positive effects of $Act_1$ is incompatible with and prevents $Act_2$.

The goal of our research regarding the computation of the *Prevents/2* relation between actions focuses on the application of the deduction axiom schema reflecting clause 3 of the definition, which in turn depends on the incompatibility and the prevention relations between a proposition and an action:

[ Action($A_1$) ∧ Action($A_2$) ∧
  ∃p (p∈PosEffects($A_1$) ∧ Incompatible(p, $A_2$) ∧ Prevents(p, $A_2$)) ] ⇒ Prevents($A_1$, $A_2$)

*Proposition Prop prevents action Act*

4. Proposition *Prop* prevents action *Act* if *Prop* prevents at least one of the preconditions of *Act*.

[ FluentProposition(P) ∧ Action(A) ∧
  ∃q (q∈Precond(A) ∧ Prevents(P, q))   ] ⇒ Prevents(P, A)

*Action Act prevents proposition Prop*

5. Action *Act* prevents fluent proposition *Prop* if at least one of the positive effects of *Act* is incompatible with and prevents proposition *Prop*.

[ Action(A) ∧ FluentProposition(P) ∧
  ∃q (q∈PosEffects(A) ∧ Incompatible(q, P) ∧ Prevents(q, P)) ] ⇒ Prevents(A, P)

The relation *Prevents/2* between propositions is learned by the agent through observation. The other cases of the relation, involving two actions or an action and a proposition, are derived using the presented deduction axiom schemata.

For explaining the fundamental ideas behind the algorithm for computing the set of all pairs of propositions such that one prevents the other, it is important to notice that there are three classes of prevention:

(i) Preventions independent of the actor of the observed behaviour and its capabilities and goals, which appear due to the nature of the world (e.g., Prevents(ActorAge(20), ActorAge(10)));
(ii) Preventions independent of the goals of the actor of the observed behaviour, which appear because the actor, using its capabilities, cannot achieve certain states after having achieved some others (e.g., Prevents(SmashedEgg(E01), IntactEgg(E01))); and
(iii) Preventions that appear because the actor's goals lead it to behave in such a way that, after a certain moment in time, there are propositions that will not happen again (e.g., Prevents(On(C, B), On(A, C))).

The algorithm has two stages. The first stage identifies preventions of the three classes. The second stage of the algorithm receives the pairs of propositions <$P_1$, $P_2$> from the first stage and removes those of the third class.

The main idea of the first stage consists of computing the pairs of propositions <$P_1$, $P_2$>, such that $P_2$ never occurs after $P_1$ has occurred.

The main idea of the second stage consists of removing those pairs of propositions <$P_1$, $P_2$>, such that it is possible to generate an action plan that achieves a state in which $P_2$ is true, from a state in which $P_1$ is true. Those action plans can only be made of the actions available to the actor of the observed behaviour. If the actor could have behaved in such a way to achieve $P_2$ after it has achieved $P_1$, then $P_1$ does not actually prevent $P_2$.

### *Proposition definition*

It may happen that a certain domain concept can be defined as a composition of other domain concepts. For example, in the Blocks World, the concept that several blocks form a stack could be expressed in terms of the concept that a block is on top of another block:

Defining(Stacked([A, B, C]), {On(A, B), On(B, C), Clear(A)})

The predicate *Defining/2* can be used to express such definitions. Defining(Prop, PropSet) means the proposition *Prop* is defined as the conjunction of all propositions in the set *PropSet*.

Agents can learn definitions as the one exemplified above if they happen to have the necessary set of sensors. It is possible to learn the mentioned definition if the agent has a sensor for facts of the form On(block, place), Clear(place), and Stacked(stack). Those sensors will acquire all the propositions On(A, B), On(B, C), Clear(A), and Stacked([A, B, C]) in every state in which A, B, and C are stacked.

The algorithm we have used to learn definitions (i.e., instances of the *Defining/2* relation) relies on the following explanation.

Imagine that the agent always observes propositions $P_2$ and $P_1$ every time proposition $P_3$ is observed. It could be assumed that $P_3$ is defined in terms of $P_2$ and $P_1$, $P_3 \equiv P_1 \wedge P_2$. However, it could also be assumed that $P_2 \equiv P_1 \wedge P_3$ or $P_1 \equiv P_2 \wedge P_3$. Nevertheless, if $P_1$, but not $P_3$, were observed in a certain state, we would know that the possible definition $P_1 \equiv P_2 \wedge P_3$ is not possible. And, if $P_2$, but not $P_3$, were observed in a certain state, then $P_2 \equiv P_1 \wedge P_3$ is not the case either.

In the first stage of the algorithm, for all fluent propositions *P*, it builds the set of pairs <P, CoOccurringProps> in which *CoOccurringProps* is the set of propositions that always co-occur with *P*. The pairs <P, CoOccurringProps> for which CoOccurringProps is the empty set are discarded.

In the second stage, the algorithm removes, from each set *CoOccurringProps*, the propositions for which a definition has been found. This avoids the presence of redundant propositions in the definitions. In the resulting set of pairs <P, Definition>, *Definition* is the set of propositions whose conjunction is the definition of *P*.

## *Although/5, Although/4*

Expressions as *although*, *in spite of*, *albeit*, *despite the fact that* and *even though* are used in natural language to express that a state or event has occurred that causes some perplexity. Often, those expressions also entail a certain degree of acceptance of the perplexing occurrence because there might have been some justification for it. Such justification, if perceived, may even eliminate or at least mitigate the caused perplexity.

We propose the relations *Although/4* and *Although/5* to capture some of the meanings of such expressions as *although* and *in spite of*. As we have previously seen, those relations were both inspired in the relation *Concession* of the Rhetorical Structure Theory (RST) [Mann and Thompson 1988].

Although(PropSet$_1$, Act, State, PropSet$_2$, Rational) may informally be read as "The observed action (*Act*) causes perplexity because, although the state of the world before its execution was more favourable for the actor of the observed behaviour (as shown by *PropSet$_1$*) than the resulting state *State* (as shown by *PropSet$_2$*), the actor actually executed it. *Rational* is the justification for, in spite of all that, the action being executed".

Although/4 is used exactly with the same sense as Although/5, but it does not specify a justification for the observed action.

*Although/5* example

Although({Desired(On(A, B)), On(A, B)/S$_1$}, Move(A, B, P$_2$), S$_2$, {Desired(On(A, B)), ¬(On(A, B)/S$_2$)}, MustPrecede(On(B, C), On(A, B)):[Move(A, B, P$_2$), Move(B, P$_1$, C), Move(A, P$_2$, B)]), which has the following informal Reading "*Although, in state S$_1$, block A was on top of block B, which is a desired state, the actor of the observed behaviour moved the block A from block B to position P$_2$ implying that A ceased to be on top of B, in state S$_2$. However, the actor did that because, if their final goal is to be achieved, block B must be placed on top of block C, before block A can be placed on top of B. The executed action, moving A from B to P$_2$, was the first action of the shortest sequence of actions that* *allowed the actor to fulfil this principle: [Move(A, B, $P_2$), Move(B, $P_1$, C), Move(A, $P_2$, B)]*".

*Although/4* example:

Although({MustPrecede(On(B, C), On(A, B)), ¬(On(A, B)/$S_0$)}, Move(A, C, B), $S_1$, {MustPrecede(On(B, C), On(A, B)), On(A, B)/$S_1$, ¬∃s (s < $S_1$ ∧ On(B, C)/s))}), which has the following informal reading: *"Although block A can only be placed on top of B after B is on top of C, and block A was not on top of block B in state $S_0$, the actor of the observed behaviour moved A from C to B leading to a state in which block A is on top of B before B is on top of C"*.

From the two examples, it is obvious that both relations *Although/5* and *Although/4* do not state general properties of the whole class of behaviours. Instead, they apply only to specific behaviours.

One of the challenges for the complete and accurate formalization of the *Although* relations consists of determining the cause of perplexity that its natural language usage entails. The causes of perplexity, in the two presented examples, are superficially different however they share a similar principle.

In the first example, there is a desired state that is fulfilled. The cause of perplexity lies in the fact that the actor of the observed behaviour destroys that desired state. That is, the actor's action removed it from an ideality.

In the second example, there is a principle according to which, block B must be on top of block C before A can be placed on top of B. In a certain state, this principle was not at stake because block A was not on top of B. Nevertheless, the actor of the observed behaviour moved A to the top of B, before B is on top of C. Once more, the actor was closer to the fulfilment of the ideality principle before it executed that action and further away from it after it executed the action. That is the cause of perplexity.

In both cases, the perplexity is caused by the execution of an action that moves the actor of the observed behaviour further away from fulfilling a certain ideality principle (e.g., Desired(On(A, B)), in the first example, and MustPrecede(On(B, C), On(A, B)), in the second example).

Finding out that the effects of an executed action move the actor of the observed behaviour further away from the fulfilment of an ideality principle involves a kind of reasoning as used in deontic systems ([Hilpinen 1981][Gabbay *et al* 2013]). The approach we propose for an agent to infer the causes of perplexity that feed the relationships of the family *Although*, through observation, consists of handling desired states (e.g., those resulting from relations *Desired/1* and *Undesired/1*) and all other relations entailing behavioural principles (e.g., *MustPrecede/2* and *Mandatory/1*) as ideality principles, in the sense that they should be fulfilled by the actor during its behaviour.

The other challenge, only pertaining *Although/5*, consists of finding the reason that justifies the action that moved the actor further away from fulfilling an ideality principle. Our approach consists of determining if the executed action (the action that moved the actor further way from fulfilling the ideality principle) was enacted at the service of another ideality principle, at least as important as the endangered one. For this approach, it is necessary to *(i)* define an order relation among principles; and *(ii)* determining if the executed action was enacted at the service of another principle. The first example illustrates the application of the described approach.

The action Move(A, B, $P_2$) moves the behaviour's actor further away from the fulfilment of the ideality principle Desired(On(A, B)). However, the same action was the first of a shortest sequence of actions that lead the actor to the fulfilment of the ideality principle MustPrecede(On(B, C), On(A, B)), which is more important than the threatened one (i.e., Desired(On(A, B))). We assume that an action was enacted at the service of an ideality principle if that action is the first of an optimal action sequence that leads to a state in which the principle is fulfilled. If the executed action does not belong to such an optimal action sequence, we assume that the actor could have behaved differently. That is, it could have maybe avoided executing the action that threatened the fulfilment of the ideality principle.

**Rigorous definition**

The axiomatic rigorous definition of the relations *Although/5* and *Although/4* relies on the identification of ideality principles and on the definition of relations aimed at determining their satisfaction degree in a given state, which was inspired in deontic concepts [Hilpinen 1981][Gabbay *et al* 2013].

*Ideality principles*

The proposed ideality principles do not specify a moment for being satisfied. A certain behaviour is compliant with an ideality principle if the principle is fulfilled by the behaviour. We propose the following ideality principles: Desired(P), Undesired(P), Mandatory(P) and MustPrecede($P_1$, $P_2$).

Desired(P): Proposition P is one of the goals of the actor of the observed behaviour. Desired(P) is an ideality principle, in the sense that, ideally, the observed behaviour should include a state in which P holds.

Undesired(P): Proposition *P* is undesired if it prevents the occurrence of a desired entity. Undesired(P) is an ideality principle in the sense that, ideally, the observed behaviour should not include a state in which the proposition holds.

Mandatory(P): Proposition *P* is mandatory if it is not one of the goals of the behaviour's actor but the observed behaviour should include a state in which *P* holds, if the actor's goals are to be achieved. Mandatory(P) is an ideality principle because, ideally, the observed behaviour should include a state in which *P* holds.

MustPrecede($P_1$, $P_2$): The actor of the observed behaviour will only be capable of achieving their goals if *(i)* in the initial state, both *$P_1$* and *$P_2$* are true; or if *(ii)* proposition *$P_1$* is true before the proposition *$P_2$* is true.

MustPrecede($P_1$, $P_2$) sets an ideality because, in the sense that if *$P_1$* and *$P_2$* are not both true in the initial state, the observed behaviour must be such that *$P_1$* is true before *$P_2$* is true.

Before introducing the relationships representing the satisfaction degree of ideality principles in a given state, it is necessary to present the notion of tagged propositions.

A tagged proposition is a pair P/S such that *P* is a proposition holding in state *S* (e.g., On(A, B)/$S_3$ – bock A is on top of block B, in state *$S_3$*) or P is a proposition independent of a particular state (e.g., Desired(On(A, B)) – the actor of the observed behaviour wants to achieve a state *in which block A is on top of block B*). If *TP* is a tagged proposition, then any logic or quantified proposition involving *TP* is also a tagged proposition, for instance ¬(P/S), which means that *P* does not hold in state *S*.

*Ideality principle satisfaction relations*

We proposed the definition of five relations expressing different degrees of ideality principle satisfaction: fulfilled, in an indifferent state, not fulfilled, not violated, and prevented.

The following represents the sequence of satisfaction relations from the one that represents a highest degree of satisfaction (i.e., fulfilled), to the one that captures a situation in which the ideality principle cannot be satisfied anymore (i.e., prevented):

*Fulfilled > Indifferent_state > Not_fulfilled > Prevented*

Additionally, we have defined the relation *Not_violated* which is true of a certain principle, in case the principle is fulfilled, or in a state indifferent to the principle, or not fulfilled (in the sense of the relation *Not_fulfilled*), although not prevented from being fulfilled.

Fulfilled(Principle, State, PropSet): The ideality principle *Principle* is satisfied in state *State*. *PropSet* is the smallest set of tagged propositions, including *Principle*, holding in state *State* or before, which certify the satisfaction of the principle.

Not_fulfilled(Principle, State, PropSet): The ideality principle Principle is not satisfied in state State, Sate is not indifferent to the principle and the principle was not prevented from being achieved before State. PropSet is the smallest set of tagged propositions, including Principle, true in state State or before it, that certify that the principle is not satisfied (in the sense of the relation *Not_fulfilled/3*). Not_fulfilled(Principle, State, PropSet) does not mean the same as ¬Fulfilled(Principle, State, PropSet) because, in states in which the principle has been violated or in states indifferent to the principle, ¬Fulfilled(Principle, State, PropSet) is true, but Not_fulfilled(Principle, State, PropSet) is not. However, whenever Not_fulfilled(Principle, State, PropSet) is true, ¬Fulfilled(Principle, State, PropSet) is also true:

Not_fulfilled(Principle, State, PropSet) $\Rightarrow$ ¬Fulfilled(Principle, State, PropSet)

Indifferent_state(Principle, State, PropSet): The state *State* is indifferent with respect to the ideality principle *Principle*. *PropSet* is the smallest set of tagged propositions, including *Principle*, true in state *State* or before it, which certify that the state is indifferent with respect to the principle. *Indifferent_state/3* is defined only for the principle MustPrecede($P_1$, $P_2$). State *S* is indifferent with respect to the precedence principle MustPrecede($P_1$, $P_2$) if $P_2$ is not true in *S*.

Not_violated(Principle, State, PropSet): The ideality principle *Principle* is not violated in the state *State*. *PropSet* is the smallest set of tagged propositions including *Principle*, true in the state *State* or before it, which certify that the principle is not violated. A principle is not violated in the state S if the principle is satisfied in *S*, if S is indifferent to the principle, or if the principle is not satisfied in *S*, although not prevented from being satisfied (in the sense of the relation *Not_fulfilled*).

Prevented(Principle, State, PropSet): The ideality principle *Principle* is not satisfied in the state *State*, and it is prevented from being satisfied in any state occurring after *State*. *PropSet* is the smallest set of tagged propositions including *Principle*, which certify that the principle was prevented from being satisfied from the state *State* onwards. For all that matters, even though the observed behaviour might not be complete, the principle is doomed to be violated because it was prevented from being satisfied.

Both relations *Although/5* and *Although/4* identify actions, executed by the actor of the observed behaviour, that move it further away from the satisfaction of a certain ideality principle. Additionally, *Although/5* identifies a justification for the execution of that action. In spite the action has moved the actor further away from the satisfaction of an ideality principle, the action was the first of an optimal action sequence that leads the actor from the state in which it was executed to the state in which an ideality principle, at least as important as the threatened one, is satisfied. This means that *Although/5* can be derived from *Although/4*.

To formalize the deduction axioms of relations *Although/4* and *Although/5*, we first present some prior explanations and notational conventions.

The relational operator *IdPr* (*Ideality Principle*) is true of ideality principles. *IdPr(Pr)* means that the proposition *Pr* represents an ideality principle in the sense that ideally *Pr* should be satisfied in one of the states of the observed behaviour.

It is necessary to define an order relation among ideality principles. If *Pr$_1$* and *Pr$_2$* are two ideality principles, *Pr$_1$* ≤ *Pr$_2$* means that, for the actor of the observed behaviour, *Pr$_2$* is as important as or more important than *Pr$_1$*.

*Pr$_t$* and *Pr$_j$* are ideality principles. *Pr$_t$* (*Threatened Principle*) was used to represent the ideality principle from which the actor of the observed behaviour was moved further away by the action *A* it executed. *Pr$_j$* (*Justifying Principle*) is the ideality principle that justifies the execution of the apparently perplexing action *A*.

*{Pr$_t$}* ∪ *PSet* is the smallest set of tagged propositions including *Pr$_t$*, true in the state in which *A* was executed or before it, which certify a certain degree of satisfaction of *Pr$_t$* in the state in which *A* was executed.

*Dev* (*Deviation*) is the smallest set of tagged propositions including *Pr$_t$*, true in the state immediately after A was executed (*S*) or before it, which certify a certain degree of satisfaction of *Pr$_t$*, in state *S*. The degree of satisfaction of *Pr$_t$*, in *S*, is less than its degree of satisfaction in the state in which *A* was executed.

It is also necessary to define a temporal relation between states of an observed behaviour. If S$_1$ and S$_2$ are state identifiers (of the same behaviour), S$_1$ < S$_2$ means that S$_1$ occurred earlier than S$_2$; and S$_1$ ≤ S$_2$ means that S$_1$ occurred earlier than or is the same as S$_2$.

Finally, the symbolic structure [X|Seq] represents a sequence that starts with X and proceeds with the possibly empty sequence Seq. That is, X is the first element of [X|Seq] and Seq is its rest.

Deduction axiom schema for the relation *Although/5*:

( Although({Pr$_t$}∪PSet, A, S$_2$, Dev) ∧
  NextState(S$_1$, A, S$_2$) ∧
  IdPr(Pr$_j$) ∧ Pr$_t$ ≤ Pr$_j$ ∧
  ¬∃$_{ps}$ Fulfilled(Pr$_j$, S$_1$, ps) ∧
  OptimumASequence(Pr$_j$, S$_1$, [A|ASeq]) ∧
  ObservedASequence(Pr$_j$, S$_1$, [A|ASeq])  ) ⇒
          Although({Pr$_t$}∪PSet, A, S, Dev, Pr$_j$:[A|ASeq])

The existential variable *ps* (*proposition set*) represents the set of tagged propositions including *Pr$_j$*, holding in state S$_1$ or before it, which certify that *Pr$_j$* would be fulfilled in state S$_1$ (if it actually were fulfilled, which it is not). The existential quantification over sets of propositions does not constitute a computational problem because what counts is whether or not the principle is fulfilled. The set of propositions certifying that the principle is fulfilled is known *a priori* by the used axiom.

*OptimumASequence(Pr, S, Seq)* means that *Seq* is an optimal action sequence that, if executed from state *S*, will result in a state in which the ideality principle *Pr* is satisfied.

*ObservedASequence(Pr, S, Seq)* means that *Seq* is the action sequence, actually executed by the actor of the observed behaviour, from state *S*, resulting in a state in which *Pr* is satisfied for the first time after state *S*.

The following deduction axiom schemata exemplify three cases in which the relation *Although/4* can be inferred because the actor of the observed behaviour executes an action that moves it further away from the satisfaction of an ideality principle:

( IdPr(Pr) ∧
  Fulfilled(Pr, $S_1$, PSet) ∧ NextState($S_1$, A, $S_2$) ∧
  Not_fulfilled(Pr, $S_2$, Dev) ) ⇒ Although(PSet, A, $S_2$, Dev)

( IdPr(Pr) ∧
  Indifferent_state(Pr, $S_1$, PSet) ∧ NextState($S_1$, A, $S_2$) ∧
  Not_fulfilled(Pr, $S_2$, Dev) ) ⇒ Although(PSet, A, $S_2$, Dev)

( IdPr(Pr) ∧
  Not_violated(Pr, $S_1$, PSet) ∧ NextState($S_1$, A, $S_2$) ∧
  Prevented(Pr, $S_2$, Dev) ) ⇒ Although(PSet, A, $S_2$, Dev)

The first axiom applies to the cases where the perplexity is caused by the execution of an action (executed in state $S_1$) responsible for the transition from a state in which a certain ideality principle is satisfied to a state in which the same ideality principle is not satisfied (in the sense of relation *Not_fulfilled*). *PSet* is the smallest set of tagged propositions including the principle *Pr*, which certify the satisfaction of *Pr* in state $S_1$; and *Dev* (from deviation) is the smallest set of tagged propositions including *Pr* that certify that *Pr* is not satisfied in state $S_2$.

The second axiom applies to cases in which the perplexity is caused by the execution of an action responsible for the transition from a state indifferent to the ideality principle *Pr* to a state in which *Pr* is not satisfied (in the sense of relation *Not_fulfilled*).

The third axiom applies to cases in which the perplexity is caused by the execution of an action responsible for the transition from a state in which the ideality principle *Pr* is not violated (implying it is not prevented from being fulfilled) to another state in which *Pr* is prevented from being satisfied.

All the exemplified deduction axiom schemata rely on the determination of the degree to which a certain ideality principle is satisfied. We have defined five degrees of satisfaction of ideality principles: satisfied in state *S* (*Fulfilled/3*), in a state indifferent to the principle (*Indifferent_state/3*), not satisfied in state *S* (*Not_fulfilled/3*), not violated in state *S* (*Not_violated/3*) and prevented, in state *S*, from being satisfied in any state occurring after *S* (*Prevented/3*). Following, we present examples of deduction axiom schemata for deriving the state of satisfaction of a principle for the five defined degrees. The formalization of those axioms uses the relation *InitialState/1* that identifies the initial state of the considered behaviour, and the function *StateProps/1* that returns the set of propositions o the specified state.

The antecedent of the implication of each deduction axiom used to infer the degree to which the specified ideality principle is satisfied includes the corresponding ideality principle. That is, the axioms may only be used to infer the degree to which an already existing principle is satisfied. The axioms cannot be used to infer ideality principles of which the observing agent is not yet aware. This means that the well-formedness of the considered ideality principle does not have to be ensured by the axiom itself. For instance, the axioms regarding the degree to which the principle MustPrecede($P_1$, $P_2$) is satisfied does not have to check whether $P_1$ and $P_2$ are fluent propositions or to ascertain that $P_1$ is not a precondition of any of the actions of the action sequence leading from the state in which $P_1$ (but not $P_2$) is true to the state in which $P_2$ is also true.

*Fulfilled(Principle, State, PropsSet)*

The precedence relation MustPrecede($P_1$, $P_2$) is satisfied in the initial state, if both propositions are true in the initial state.

( MustPrecede($P_1$, $P_2$) ∧
  InitialState(S) ∧ $P_1$ ∈ StateProps(S) ∧ $P_2$ ∈ StateProps(S) ) ⇒
    Fulfilled(MustPrecede($P_1$, $P_2$), S, {MustPrecede($P_1$, $P_2$), $P_2$/S, $P_1$/S,
                                              InitialState(S)})

The precedence relation MustPrecede($P_1$, $P_2$) is satisfied in state $S_2$, in which $P_2$ is true, if $P_1$ but not $P_2$ is true in a state $S_1$ occurring before $S_2$.

( MustPrecede($P_1$, $P_2$) ∧ State($S_1$) ∧ State($S_2$) ∧
  $P_2$ ∈ StateProps($S_2$) ∧ $P_1$ ∈ StateProps($S_1$) ∧ $P_2$ ∉ StateProps($S_1$) ∧ $S_1 < S_2$ ) ⇒
    Fulfilled(MustPrecede($P_1$, $P_2$), $S_2$,
         {MustPrecede($P_1$, $P_2$), $P_2$/$S_2$, $P_1$/$S_1$, ¬($P_2$/$S_1$), $S_1$<$S_2$})

Desired(P) is satisfied in state S, if P is true in S.

( Desired(P) ∧ State(S) ∧ P∈StateProps(S)) ) ⇒
    Fulfilled(Desired(P), S, {Desired(P), P/S})

*Indifferent_state(Principle, State, PropsSet)*

A state S, other than the initial one, is indifferent to the precedence principle MustPrecede($P_1$, $P_2$), if $P_2$ does not hold in S.

( MustPrecede($P_1$, $P_2$) ∧ State(S) ∧ ¬InitialState(S) ∧ $P_2$ ∉ StateProps(S) ) ⇒
    Indifferent_state(MustPrecede($P_1$, $P_2$), S,
         {MustPrecede($P_1$, $P_2$), ¬InitialState(S), ¬($P_2$/S)})

The initial state S is indifferent to the precedence principle MustPrecede($P_1$, $P_2$), if neither $P_1$ nor $P_2$ hold in S.

( MustPrecede($P_1$, $P_2$) ∧ InitialState(S) ∧ $P_1$ ∉ StateProps(S) ∧ $P_2$ ∉ StateProps(S) ) ⇒
    Indifferent_state(MustPrecede($P_1$, $P_2$), S,
         {MustPrecede($P_1$, $P_2$), InitialState(S), ¬($P_1$/S), ¬($P_2$/S)})

*Not_fulfilled(Principle, State, PropsSet)*

We have decided to define the relation *Not_fulfilled/3*, which states that a principle is not fulfilled but it has not been prevented from being fulfilled. For the accurate definition of the relation *Not_fulfilled/3*, it is convenient to use the relation *PreventedProp/2*, such that PreventedProp(P, S) means that, in state S, the proposition P was prevented from occurring in any state occurring after S. The following is the deduction axiom for relationships of the relation *PreventedProp/2*, assuming that P is a fluent proposition:

∀q [(FluentProposition(q) ∧ State(S) ∧ q ∈ StateProps(S) ∧ Prevents(q, P)) ⇒
                    PreventedProp(P, S)]

Notice that the above axiom implies that P was prevented in state S, but it does not preclude the possibility that P has been prevented in a state before S. Notice also that PreventedProp(P, S) corresponds to the tagged proposition PreventedProp(P)/S, with a similar deduction schema:

∀q (FluentProposition(q) ∧ State(S) ∧ q/S ∧ Prevents(q, P)) ⇒
                    PreventedProp(P)/S

In [Botelho et al 2019], we have decided that the precedence relation MustPrecede($P_1$, $P_2$) is trivially satisfied if $P_1$ and $P_2$ are both true in the initial state of the considered behaviour. The following axiom schema presents the conditions under which a precedence relation is not fulfilled (in the sense of the relation *Not_fulfilled/3*) in a state other than the initial one.

[ MustPrecede($P_1$, $P_2$) ∧ State(S) ∧ ¬InitialState(S) ∧ $P_2$∈ StateProps(S) ∧
  ¬∃$t_1$ (State($t_1$) ∧ $t_1$ < S ∧ $P_1$ ∈ StateProps($t_1$)) ∧
  ¬∃$t_2$ (State($t_2$) ∧ $t_2$ < S ∧ PreventedProp($P_1$, $t_2$)) ] ⇒
        Not_fulfilled(MustPrecede($P_1$, $P_2$), S,
                {MustPrecede($P_1$, $P_2$), ¬InitialState(S), $P_2$/S,
                ¬∃$t_1$ ($t_1$ < S ∧ $P_1$/$t_1$), ¬∃$t_2$ ($t_2$ < S ∧ PreventedProp($P_1$)/$t_2$) })

The tagged proposition ¬∃$t_1$ ($t_1$ < S ∧ $P_1$/$t_1$) means that there is no state $t_1$, occurring before state *S*, in which proposition $P_1$ holds, and the tagged proposition ¬∃$t_2$ ($t_2$ < S ∧ PreventedProp($P_1$)/$t_2$) means that there is no state $t_2$ prior to S in which $P_1$ has been prevented from occurring in states occurring after $t_2$ (of the same considered behaviour instance). For the simpler case in which the precedence relation would be satisfied if the two propositions were true in the initial state, the deduction axiom schemata are much simpler:

( MustPrecede($P_1$, $P_2$) ∧ InitialState(S) ∧ $P_2$∈ StateProps(S) ∧ $P_1$∉ StateProps(S) )
    ⇒ Not_fulfilled(MustPrecede($P_1$, $P_2$), S,
        {MustPrecede($P_1$, $P_2$), InitialState(S), $P_2$/S, ¬($P_1$/S)})

( MustPrecede($P_1$, $P_2$) ∧ InitialState(S) ∧ $P_1$∈ StateProps(S) ∧ $P_2$∉ StateProps(S) )
    ⇒ Not_fulfilled(MustPrecede($P_1$, $P_2$), S,
        {MustPrecede($P_1$, $P_2$), InitialState(S), $P_1$/S, ¬($P_2$/S)})

in which the tagged propositions of the form ¬(P/S) mean that proposition *P* is not the case in state *S*.

The next deduction axiom schema addresses the case in which the satisfaction degree of the ideality principle Desired(P) is not fulfilled (implying it is also not prevented).

[ Desired(P) ∧ State(S) ∧ P∉StateProps(S) ∧ ¬∃t (State(t) ∧ t ≤ S ∧ PreventedProp(P, t) )]
    ⇒ Not_fulfilled(Desired(P), S, {Desired(P), ¬(P/S),
            ¬∃t (State(t) ∧ t ≤ S ∧ PreventedProp(P)/t)  })

*Not_violated(Principle, State, PropsSet)*

Fulfilled principles, principles in states indifferent to them, and principles not fulfilled but also not prevented are all not violated.

Indifferent_state(Pr, S, PSet) ⇒Not_violated(Pr, S, PSet)

Fulfilled(Pr, S, PSet) ⇒ Not_violated(Pr, S, PSet)

Not_fulfilled(Pr, S, PSet) ⇒Not_violated(Pr, S, PSet)

*Prevented(Principle, State, PropsSet)*

Desired($P_2$) is prevented in state S from ever being fulfilled in states to come, if $P_2$ does not hold in S, but there is a proposition holding in S that prevents $P_2$ from occurring after S.

( Desired($P_2$) ∧ Prevents($P_1$,$P_2$) ∧ State(S) ∧ $P_2$∉ StateProps(S) ∧ $P_1$∈ StateProps(S) )
        ⇒Prevented(Desired($P_2$), S, {Desired($P_2$), Prevents($P_1$, $P_2$), $P_1$/S, ¬($P_2$/S)})

The algorithms for inferring propositions of the relations *Although/4* and *Although/5* directly apply the deduction axioms defined for those relations. Actually, each axiom corresponds to a special case of the algorithm.

To derive a fact of the relation *Although/4*, it is necessary to determine the degree to which a certain ideality principle is satisfied.

In the case of the first axiom of the relation *Although/4*, for example, the algorithm checks if there is an ideality principle that is satisfied (*Fulfilled/3*) in the state *S*, before the action is executed, and not satisfied (*Not_fulfilled/3*) in the next state.

We have proposed deduction axioms for each defined degree of satisfaction of ideality principles (whenever applicable). In each case, the axiom provides four pieces of information: the deduced degree of satisfaction, the principle whose degree of satisfaction was deduced, the state of the considered behaviour in which the degree of satisfaction was deduced, and the smallest set of tagged propositions, true in that state or before it, that certify the deduced satisfaction degree (including the ideality principle).

It may be seen that each axiom for deriving the degree to which a certain ideality principle is satisfied knows the smallest set of tagged propositions that certify that satisfaction degree. The set is written along with the axiom. Likewise, the algorithm implementing each such axiom also possesses the same information.

The deduction axioms for the relation *Although/4* also depend on the relation *IdPr/1*, which keeps the considered set of ideality principles. *IdPr/1* is provided as input (i.e., pre-programmed) to the used algorithms.

Following its presented axiom, the deduction of propositions of the relation *Although/5* depends of the deduction of propositions of the relation *Although/4*, of the relation of all considered ideality principles (*IdPr/1*), of the order relation reflecting the relative importance of ideality principles to the actor of the observed behaviour, of the determination that a principle is not satisfied, of the determination of the optimal action sequences that satisfy a principle starting from a given state, and of the determination of the action sequence actually executed by the actor of the observed behaviour to satisfy the considered principle, starting from the same state.

( Although({$Pr_t$}∪PSet, A, $S_2$, Dev) ∧
  NextState($S_1$, A, $S_2$) ∧
  IdPr($Pr_j$) ∧ $Pr_t$ ≤ $Pr_j$ ∧
  ¬∃$_{ps}$ Fulfilled($Pr_j$, $S_1$, ps) ∧
  OptimumASequence($Pr_j$, $S_1$, [A|ASeq]) ∧
  ObservedASequence($Pr_j$, $S_1$, [A|ASeq])  ) ⇒
        Although({$Pr_t$}∪PSet, A, S, Dev, $Pr_j$:[A|ASeq])

Of all of these, we have only to discuss the order relation among ideality principles, and the determination of the optimal and actual action sequences leading to the satisfaction of a certain principle.

In the proof of concept demonstration (section 3.2), the order relation among ideality principles was arbitrarily defined (in the sense that it was not derived of more fundamental principles).

We have used an adaptation of an optimum path algorithm for the determination of the optimum action sequence that satisfies a given ideality principle, starting from a specified state. The algorithm is similar to a planning algorithm limited to producing action sequences not longer than the number of actions executed in the considered behaviour by its actor, since the considered state. The major adaptation of the algorithm consisted of replacing the usual notion of goal satisfaction with the set of axioms for determining if the considered ideality principle is satisfied.

The algorithm that determines the actual action sequence executed by the actor of the observed behaviour, from the considered state until the first state in which the considered ideality principle is satisfied, orderly traverses the observed states and applies the axioms that determine the degree of satisfaction of the ideality principle to each visited state.

### *Background(PropsSet, Relationship)*

The *Background/2* relation may be used to specify the background domain knowledge used by an agent to infer a given relationship between domain entities. *Background/2* is presented as a possibility. However, we preserve our strong determination of avoiding using background domain knowledge.

In the following example, the proposition $\forall x \forall y\ On(x, y) \Rightarrow \neg Clear(y)$ is the domain background knowledge used by the agent to infer that $On(A, B)$ is incompatible with $Clear(B)$. The conclusion $Incompatible(On(A, B), Clear(B))$ is derived from the domain background knowledge $\forall x \forall y\ On(x, y) \Rightarrow \neg Clear(y)$, from the hypothesis $On(A, B)$, and from the relevant clause of the definition of the relation *incompatibility* (Def#2).

The definition of incompatibility between propositions consists of the following two clauses:

Def#1. Two propositions are incompatible if one is the negation of the other

Def#2. Two propositions are incompatible if, although neither is the negation of the other, it is possible to infer the negation of one of them *(i)* from the other, and possibly *(ii)* from background or learned domain knowledge.

Following clause Def#2, it is possible to conclude that $On(A, B)$ is incompatible with $Clear(B)$:

1. $\forall x \forall y\ On(x, y) \Rightarrow \neg Clear(y)$      $\Delta$ (background knowledge)
2. $On(A, B)$      hypothesis
3. $On(A, B) \Rightarrow \neg Clear(B)$      1 2×Universal Instantiation
4. $\neg Clear(B)$      2, 3 Modus Ponens
5. $Incompatible(On(A, B), Clear(B))$      Def#2 and the inference of 4 from 1, 2

That is, $\forall x \forall y\ On(x, y) \Rightarrow \neg Clear(y)$ is the background knowledge used by the agent to infer $Incompatible(On(A, B), Clear(B))$ from $On(A, B)$.

In this article, the agent does not use background knowledge to infer the relationships describing the observed behaviour, its actor and environment. However, if it happens to be necessary or otherwise desired, the *Background/2* relation specifies the observing agent's prior background knowledge used to infer the considered relationship.

Let's assume the agent used background domain knowledge to derive a certain proposition Q. The following statements present the deductive process used to infer instances of the relation *Background/2*.

Let $\Delta$ be the set of beliefs the agent possesses before observing the considered behaviour.

Let $\Omega$ be the set of beliefs the agent acquires through observation, from the moment where the observation started until the instant of time immediately before the moment in which the agent started inferring Q. $\Omega$ must include propositions that could not have been derived from $\Delta$, using the agent's deductive apparatus. To express this condition, consider the function *Closure* that represents the closure of a given set of propositions under the agent's deductive apparatus. $(\Omega - Closure(\Delta)) \neq \varnothing$ means that there are propositions in $\Omega$ that cannot be derived from $\Delta$.

Let Ψ be the set of propositions actually used by the agent to derive Q, using its deductive system, Ψ ⊢ Q, with the restriction that each proposition in Ψ is an hypothesis, belongs to Δ, or belongs to Ω, but at least one of them belongs to Δ (Δ ∩ Ψ ≠ ∅).

In the referred conditions, the following deduction rule can be used to infer instances of the *Background/2* relation:

$$\frac{\Psi \vdash Q \quad \Psi \subseteq (\Delta \cup \Omega) \quad (\Omega - \text{Closure}(\Delta)) \neq \emptyset \quad (\Delta \cap \Psi) \neq \emptyset}{\text{Background}((\Delta \cap \Psi), Q)}$$

Using this inference rule and the deduction presented before, it is possible to derive Background({∀x∀y On(x, y) ⇒ ¬Clear(y)}, Incompatible(On(A, B), Clear(B))) (assuming the agent's deduction system includes the relevant clause of the Incompatible relation).

## 3.2 Proof of concept and discussion

The definitions and algorithms presented in sections 2 and 3.1 were demonstrated in a simple scenario of the Blocks World. Given their independence of the domain, the presented definitions and algorithms could have been demonstrated in other scenarios.

### 3.2.1 Blocks World scenario

In this version of the Blocks World, a robot moves blocks around until they are all stacked on top of each other. Thus, the robot is the actor of the observed behaviour.

The place where the blocks will be stacked is irrelevant. There are always three blocks – A, B and C – and four spaces on a table – $P_1$, $P_2$, $P_3$ and $P_4$. Each block may be placed on one of the table positions or on top of any other block. The predicate *Place/1* is used to hold the places where blocks may be located. The final stack must contain the block B on top of the block C and the block A on top of the block B.

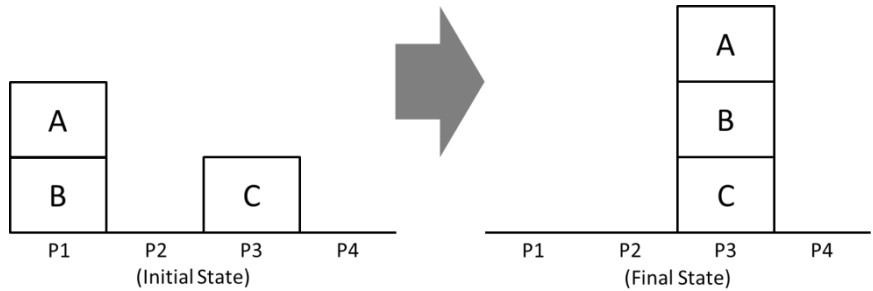

**Figure 1 – Example of a Blocks World problem**

Two blocks cannot be placed on the same position therefore a block can only be moved to a place if the place is clear. Predicate *Clear/1* is true of clear places. The robot can only move a block at a time, and only if the block to be moved does not have any other block on top of it. The predicate *On/2* is used to specify the relation between a block and the place on which it is positioned. On(block, place) means that the specified block is positioned on the specified place.

The states of the world contain only positive instances of the *Clear/1* and *On/2* predicates, which are accessible to the observing agent. The robot's only action, Move(block, from, to), moves the specified block, from the specified place (from), to the specified place (to).

We have made some simplifications in the demonstration to be described. These simplifications will be relaxed in future experiments. First, we assume that the observing agent recognizes each time a final state is reached; final states are always successful.

According to the second simplification, each action produces always the same effects. Conditional and imperfect actions were not considered.

Third, we have also assumed that the observing agent has perfect access to all relevant aspects of the world.

All initial configurations (i.e., 120) were automatically generated and the robot had to stack the blocks in each of them, which represents 120 instances of the same class of behaviours. The states and the actions of each behaviour were recorded on a file. The 120 files were processed by the defined algorithms, which acquired the defined concepts and generated explanations of the robot's behaviours.

### 3.2.2 Results

In [Botelho et al 2019], we presented and demonstrated a set of relations acquired by an agent through mere observation, with no prior knowledge of the domain. The acquired relations describe the observed behaviour, its actor and environment.

In this paper, we have extended the originally proposed set of relations with several new ones, many of which were inspired in the Rhetorical Structure Theory [Mann and Thompson 1988]. This section describes the results achieved with implemented software acting on a simple version of the Blocks World.

Our algorithms have correctly discovered the new proposed concepts: Desired actions, *Undesired/1, Neutral/1*, *Incompatible/2*, *Prevents/2*, *Defining/2, Although/5* and *Although/4. Background/2* was not tested.

#### *Desired Actions: Desired(Entity) ∧ Action(Entity)*

In [Botelho et al 2019] we have defined the relation *Desired/1* just for propositions. Here, we extend the concept also for actions. The presented deduction axiom and corresponding algorithm were used to correctly derive the set of all desired actions.

| | | | |
|---|---|---|---|
| Move(A, C, B) | Move(A, $P_1$, B) | Move(A, $P_2$, B) | Move(A, $P_3$, B) |
| Move(A, $P_4$, B) | Move(B, A, C) | Move(B, $P_1$, C) | Move(B, $P_2$, C) |
| Move(B, $P_3$, C) | Move(B, $P_4$, C) | Move(B, A, $P_3$) | Move(B, A, $P_4$) |
| Move(C, A, $P_1$) | Move(C, A, $P_2$) | Move(C, A, $P_3$) | Move(C, A, $P_4$) |

Desired actions are those whose positive effects include desired propositions, unless they destroy other desired propositions or give rise to undesired propositions.

Since, in the demonstrated Blocks World, there are not undesired propositions, they do not interfere in the determination of the desired actions.

The five first presented actions are desired because they produce On(A, B), which is a desired proposition because it is one of the robot's goals. The following five next actions all give rise to On(B, C), which is also a desired proposition. Finally, all other actions (as well as Move(B, A, C), already considered) give rise to states in which Clear(A) (also a desired proposition) holds.

There are potentially other desired actions (e.g., Move(C, A, B)) but they have not even been observed.

### Undesired entities: Undesired(Entity)

In the considered Blocks World scenario, there are no undesired entities (propositions or actions). To test the presented axiom and corresponding algorithm, we have artificially injected proposition *GoalPreventingProp*, which prevents the goal On(A, B) from being achieved. The observing agent correctly learned the proposition Prevents(GoalPreventingProp, On(A, B)), in that modified scenario. Consequently, it was also capable of deriving Undesired(GoalPreventingProp).

### Neutral propositions: FluentProposition(Entity) ∧ Neutral(Entity)

The presented deduction axioms and corresponding algorithms were used to correctly derive the set of all neutral propositions (not desired nor undesired).

| Clear(B) | Clear(C) | Clear($P_1$) | Clear($P_2$) | Clear($P_3$) | Clear($P_4$) | |
|---|---|---|---|---|---|---|
| On(A, C) | On(A, $P_1$) | On(A, $P_2$) | On(A, $P_3$) | On(A, $P_4$) | On(B, A) | On(B, $P_1$) |
| On(B, $P_2$) | On(B, $P_3$) | On(B, $P_4$) | On(C, A) | On(C, B) | On(C, $P_1$) | On(C, $P_2$) |
| On(C, $P_3$) | On(C, $P_4$) | | | | | |

### Neutral actions: Action(Entity) ∧ Neutral(Entity)

The presented deduction axioms and corresponding algorithms were used to correctly derive the set of all neutral actions.

| Move(A, B, $P_1$) | Move(A, B, $P_2$) | Move(A, B, $P_3$) | Move(A, C, $P_1$) |
|---|---|---|---|
| Move(A, C, $P_2$) | Move(A, C, $P_3$) | Move(A, C, $P_4$) | Move(A, $P_1$, $P_2$) |
| Move(A, P1, $P_3$) | Move(A, P1, $P_4$) | Move(A, P2, $P_3$) | Move(A, $P_2$, $P_4$) |
| Move(A, $P_3$, $P_2$) | Move(A, $P_3$, $P_4$) | Move(A, $P_4$, $P_2$) | Move(A, $P_4$, $P_3$) |
| Move(B, C, $P_2$) | Move(B, C, $P_4$) | Move(B, $P_1$, $P_2$) | Move(B, $P_1$, $P_3$) |
| Move(B, $P_2$, $P_3$) | Move(B, $P_3$, $P_2$) | Move(B, $P_4$, $P_2$) | Move(B, $P_4$, $P_3$) |
| Move(C, B, $P_1$) | Move(C, B, $P_2$) | Move(C, B, $P_3$) | Move(C, B, $P_4$) |
| Move(C, $P_3$, $P_1$) | Move(C, $P_4$, $P_1$) | Move(C, $P_4$, $P_2$) | |

Given that there are no undesired actions, the neutral actions are limited to those that are not desired.

### Incompatible propositions: Incompatible($P_1$, $P_2$)

The incompatibility between propositions is a symmetric relation. Following, we present the first half of the incompatibility relation for propositions. The second half can be built by symmetry.

| (Clear(A), On(B,A)) | (On(A,$P_3$), On(A,C)) | (On(B,$P_3$), On(B,A)) |
|---|---|---|
| (Clear(B), On(C,B)) | (On(A,$P_3$), On(A,$P_1$)) | (On(B,$P_3$), On(B,$P_4$)) |
| (Clear(C), On(A,C)) | (On(A,$P_3$), On(A,$P_2$)) | (On(B,$P_4$), On(B,A)) |
| (Clear(C), On(B,C)) | (On(A,$P_3$), On(A,$P_4$)) | (On(C,A), Clear(A)) |
| (Clear($P_1$), On(A,$P_1$)) | (On(A,$P_3$), On(B,$P_3$)) | (On(C,A), On(A,C)) |
| (Clear($P_1$), On(C,$P_1$)) | (On(A,$P_3$), On(C,$P_3$)) | (On(C,A), On(B,A)) |
| (Clear($P_2$), On(A,$P_2$)) | (On(A,$P_4$), On(A,C)) | (On(C,A), On(C,B)) |
| (Clear($P_2$), On(B,$P_2$)) | (On(A,$P_4$), On(A,$P_1$)) | (On(C,A), On(C,$P_1$)) |
| (Clear($P_2$), On(C,$P_2$)) | (On(A,$P_4$), On(A,$P_2$)) | (On(C,A), On(C,$P_2$)) |
| (Clear($P_3$), On(A,$P_3$)) | (On(A,$P_4$), On(B,$P_4$)) | (On(C,A), On(C,$P_3$)) |
| (Clear($P_3$), On(B,$P_3$)) | (On(B,C), On(A,C)) | (On(C,A), On(C,$P_4$)) |
| (Clear($P_3$), On(C,$P_3$)) | (On(B,C), On(B,A)) | (On(C,$P_1$), On(A,$P_1$)) |
| (Clear($P_4$), On(A,$P_4$)) | (On(B,C), On(B,$P_2$)) | (On(C,$P_1$), On(C,B)) |
| (Clear($P_4$), On(B,$P_4$)) | (On(B,C), On(B,$P_3$)) | (On(C,$P_2$), On(A,$P_2$)) |
| (Clear($P_4$), On(C,$P_4$)) | (On(B,C), On(B,$P_4$)) | (On(C,$P_2$), On(B,$P_2$)) |
| (On(A,B), Clear(B)) | (On(B,C), On(C,B)) | (On(C,$P_2$), On(C,B)) |
| (On(A,B), On(A,C)) | (On(B,$P_1$), Clear($P_1$)) | (On(C,$P_2$), On(C,$P_1$)) |
| (On(A,B), On(A,$P_1$)) | (On(B,$P_1$), On(A,$P_1$)) | (On(C,$P_2$), On(C,$P_3$)) |
| (On(A,B), On(A,$P_2$)) | (On(B,$P_1$), On(B,A)) | (On(C,$P_3$), On(B,$P_3$)) |
| (On(A,B), On(A,$P_3$)) | (On(B,$P_1$), On(B,C)) | (On(C,$P_3$), On(C,B)) |
| (On(A,B), On(A,$P_4$)) | (On(B,$P_1$), On(B,$P_2$)) | (On(C,$P_3$), On(C,$P_1$)) |
| (On(A,B), On(B,A)) | (On(B,$P_1$), On(B,$P_3$)) | (On(C,$P_4$), On(A,$P_4$)) |
| (On(A,B), On(C,B)) | (On(B,$P_1$), On(B,$P_4$)) | (On(C,$P_4$), On(B,$P_4$)) |
| (On(A,$P_1$), On(A,C)) | (On(B,$P_1$), On(C,$P_1$)) | (On(C,$P_4$), On(C,B)) |
| (On(A,$P_2$), On(A,C)) | (On(B,$P_2$), On(B,A)) | (On(C,$P_4$), On(C,$P_1$)) |
| (On(A,$P_2$), On(A,$P_1$)) | (On(B,$P_2$), On(B,$P_3$)) | (On(C,$P_4$), On(C,$P_2$)) |
| (On(A,$P_2$), On(B,$P_2$)) | (On(B,$P_2$), On(B,$P_4$)) | (On(C,$P_4$), On(C,$P_3$)) |

### *Preventing relation between propositions: Prevents(Entity$_1$, Entity$_2$)*

In the described Blocks World scenario, no proposition prevents another one from occurring. To test the algorithm, we have artificially and randomly injected propositions *PreventingP* or *P*, in all states of all behaviours of the same class, staring with the initial state until the proposition *PreventingP* was injected. When *PreventingP* was injected, no other proposition was artificially added to future states of the same behaviour. This way, we have simulated the existence of prevention relationships between propositions: *PreventingP* prevents the future occurrence of *P* and *PreventingP*.

In the described circumstances, the algorithm correctly identified the following preventing relations: Prevents(PreventingP, PreventingP) and Prevents(PreventingP, P).

We performed yet another test in which, after proposition *PreventingP* has been injected in a state, it was always injected in all states thereafter until the last state (but not proposition *P*). This way, we simulated a different preventing relation. The algorithm correctly identified the relationship Prevents(PreventingP, P) (but not Prevents(PreventingP, PreventingP)), as expected.

### *Proposition definition: Defining(Prop, PropSet)*

Since the originally described demonstration scenario is not adequate for testing the algorithm proposed for discovering domain propositions that are defined from other more fundamental ones, we have artificially added the proposition Stacked([A, B, C]) in all final states of all observed behaviours (i.e., states in which On(A, B), On(B, C) and Clear(A) are true). With this addition, the observing agent correctly identified the definition

Defining(Stacked([A, B, C]), {On(A, B), On(B, C), Clear(A)}), in which A, B and C are constants denoting blocks.

### *Perplexing actions: Although/4 and Although/5*

From the newly proposed set of relations, *Although/5* and *Although/4* are the only that enable describing specific behaviours. As explained in sections 2 and 3.1, Although(PropSet$_1$, Action, State, PropSet$_2$) and Although(PropSet$_1$, Action, State, PropSet$_2$, Rational) express the observing agent's perplexity with the robot's action. In the case of *Although/5*, *Rational* provides a justification for the apparently perplexing action.

The observing agent correctly described all cases in which the robot's behaviour was perplexing. However, it is impossible and useless to show the results of all actions of all 120 observed behaviours. Instead, we describe an example behavioural instance (Figure 2) and we illustrate the explanations discovered by the observing agent observer component.

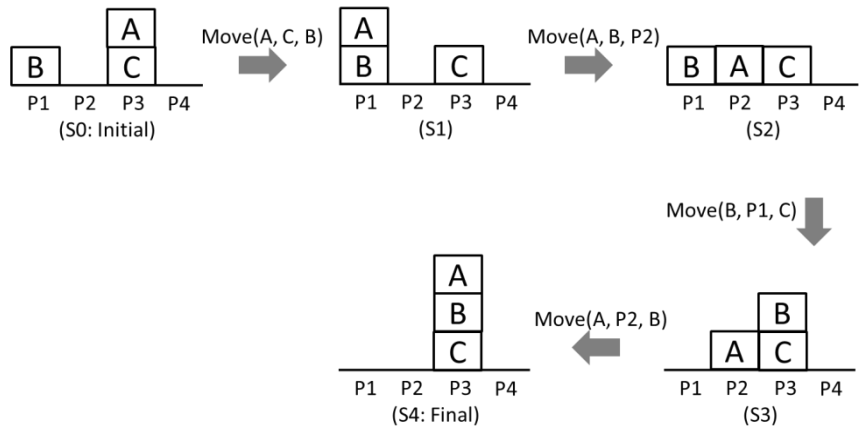

**Figure 2 – Example of a behavioural instance**

In the tests, we have only considered the ideality principles Desired(P), for propositions, and MustPrecede(P$_1$, P$_2$). The order relation of importance among principles was arbitrarily defined so that Desired(P) is less important than MustPrecede(P$_1$, P$_2$).

In this scenario, the observing agent correctly acquired the *Although/4* and *Although/5* descriptions of its behaviour.

*Although(PSet, A, S, Dev)*

Although(
    {Desired(On(A, B)), On(A, B)/S$_1$}, Move(A, B, P$_2$), S$_2$,
    {Desired(On(A, B)), ¬(On(A, B)/S$_2$)})

Although On(A, B) is a desired proposition, and On(A, B) holds in state S$_1$, the robot moved the block A from the top of B to position P$_2$, Move(A, B, P$_2$), resulting in state S$_2$, which is further away from the satisfaction of Desired(On(A, B)) because On(A, B) does not hold.

Although(
 {MustPrecede(On(B, C), On(A, B)), ¬(On(A, B)/$S_0$)}, Move(A, C, B), $S_1$,
 {MustPrecede(On(B, C), On(A, B)), On(A, B)/$S_1$, ¬∃s [s < $S_1$ ∧ On(B, C)/s]})

Although On(B, C) must precede On(A, B) and On(A, B) does not hold in the state $S_0$ (i.e., $S_0$ is indifferent to the ideality principle), the robot executed action Move(A, C, B), resulting in state $S_1$, in which A was placed on top of B before B has been placed on top of C. Thus, the new state $S_1$ is further away from the satisfaction of the ideality principle than state $S_0$.

*Although(PSet, A, S, Dev, Rational)*

Although({Desired(On(A, B)), On(A, B)/$S_1$}, Move(A, B, $P_2$), $S_2$,
 {Desired(On(A, B)), ¬(On(A, B)/$S_2$)},
 Desired(On(B, C)):[Move(A, B, $P_2$), Move(B, $P_1$, C)])

Although On(A, B) is a desired proposition and On(A, B) holds in state $S_1$, the robot executed action Move(A, B, $P_2$), leading to state $S_2$, in which On(A, B) does not hold. Nevertheless, the action Move(A, B, $P_2$) was executed at the service of the ideality principle Desired(On(B, C)), which is as important as the principle threatened by the robot's action, Desired(On(A, B)).

Although({Desired(On(A, B)), On(A, B)/$S_1$}, Move(A, B, $P_2$), $S_2$,
 {Desired(On(A, B)), ¬(On(A, B)/$S_2$)},
 MustPrecede(On(B, C), On(A, B)):[Move(A, B, $P_2$), Move(B, $P_1$, C),
                                                                            Move(A, $P_2$, B)])

Although On(A, B) is a desired proposition and On(A, B) holds in state $S_1$, the robot executed action Move(A, B, $P_2$), leading to the state $S_2$, in which On(A, B) is no longer the case. Nevertheless, the action Move(A, B, $P_2$) was executed to satisfy the precedence relationship MustPrecede(On(B, C), On(A, B)), which is an ideality principle more important than the principle threatened by the robot's action, Desired(On(A, B)).

Descriptions of the relations *Although/4* e *Although/5* constitute the basis for the agent to become astonished with the observed behaviour. Moreover, *Although/5* provides a justification for the apparently disconcerting actions.

Whereas the action Move(A, B, $P_2$) has two possible justifications: being part of the best ways to achieve On(B,C) and to fulfil the precedence relation principle, MustPrecede(On(B, C), On(A, B)), the first action executed by the robot, Move(A, C, B), is not only disconcerting because the robot becomes further away from fulfilling the ideality principle represented by the precedence relation MustPrecede(On(B, C), On(A, B)), but there is also no apparent reason for it. In fact, the first action of the robot is absolutely misplaced.

# 4 Related literature

The purpose of our research is twofold: *(i)* defining a set of concepts that may be used by an artificial agent to describe observed behaviour, its actor and environment, and *(ii)* defining and implementing the set of axiom schemata and domain-independent algorithms the agent may use to learn those concepts from scratch, relying only on observation.

We start this review with a brief reference to work on behaviour or procedure description languages. Then, we focus on literature related with more specific topics of our research, following the same structure we have used, in the introduction, to present the main contributions of our work, namely *(i)* beliefs goals and preferences, *(ii)* incompatibility between propositions and/or actions, *(iii)* perplexity, *(iv)* relation definition, and *(v)* the use of background knowledge.

Several languages have been proposed for describing behaviour (e.g., [Cerone 2019] [Piller, Vincze and Kovács 2015][Wan, Wu and Wu 2009][Kelsen, Pulvermueller and Glodt 2007][Freed and Remington 2000]). [Cerone 2019] presents a language for cognitive modelling of human behaviour and reasoning. [Piller, Vincze and Kovács 2015] and [Freed and Remington 2000] present languages that may be used to specify agent behaviour. The language proposed in [Wan, Wu and Wu 2009] is used for the specification of the behaviour of such systems as an Automatic Teller Machine (ATM). Finally, [Kelsen, Pulvermueller and Glodt 2007] presents a language for automatic system implementation. In all cases, the proposed languages are used as behaviour generation languages. For instance, using BRDL (Behaviour and Reasoning Description Language) [Cerone 2019], one can write the set of rules that specify the actions prescribed by the represented cognitive model in the specified conditions. They are not used to describe observed behaviour, thus they cannot be compared with our work. Nonetheless, some constructs could be used for describing observed behaviour. Namely, the proposal described in [Freed and Remington 2000] includes constructs useful for uncertainty, multiple tasks and concurrent execution. Our proposal does not address any of those problems.

*User modelling: beliefs, goals, preferences and emotions*. The bulk of literature more closely related with our work describes approaches that can be used to extract the model of the user from their behaviour (their beliefs, goals and preferences). Research on narrative intelligence, storytelling, and interactive digital games [Harrison *et al* 2014] [Riedl and Bulitko 2012] [Hendrikx *et al* 2013] strives to understand the user to dynamically adapt the narrative or game options to their preferences.

Computational cognitive science, computational social science, behavioural sciences, and the study of mixed team collaboration also have an interest in the problem of understanding observed action as a result of goal directed behaviour. It is important to model the way people understand observed action as if it were goal-directed [Baker, Saxe and Tenenbaum 2009]; it is important to infer individual goals in social interactions [Tauber and Steyvers 2011][Riordan *et al* 2011]; and it is also important to be capable of segmenting individual actions and inferring their causal structure [Buchsbaum *et al* 2015] (both for developing more effective artificial members of mixed collaborative teams and for modelling the way people do it).

Following, we analyse concepts and algorithms used in player modelling and behaviour understanding and compare them with our work.

Several approaches to player modelling in storytelling and digital games represent the player model as a set of preferences for different game playing styles [Ramirez and Bulitko 2014] [Thue, Bulitko and Spetch 2008], or for plot options [Yu and Riedl 2012] [Yu and Riedl 2013].

The proposed notion of preference is in line with the concept captured by the RST relation *Evaluation*. This relation could be used for representing both the preferences of the agent and those of a hypothetical audience. In fact, the agent preferences toward situations or alternative courses of action allows the agent to evaluate them and support its choices. The preferences of a hypothetical audience are the basis for plot decisions in interactive story generation systems and game style decisions in game playing settings.

In [Ramirez and Bulitko 2014] and [Thue, Bulitko and Spetch 2008], all game actions available to the player are tagged with game playing styles. When the player choses an action, if at that time the player had an available action much different from the one it chooses, the player model (his or her preferences for all types of game playing styles) is updated considering the chosen action and its nature. Once the player preferences are known, a knowledge-based approach is used to map those preferences into game management choices (e.g., players who seek out (enjoy) monster ambushes might prefer to engage in combat).

Although we have not addressed alternative behaviours, we provided the basis for a qualitative scale of preferences, through the relations *desired*, *neutral* and *undesired*, which apply both to propositions and actions. These relations, which may be learned from observation, could be used to define the relation *evaluation* inspired in the Rhetoric Structure Theory (RST). However, this would be a three-valued scale of preferences, which may be insufficient in some applications.

In the approach followed in [Yu and Riedl 2012] and [Yu and Riedl 2013] many players are asked to rate the game management options after each option is taken. These ratings are then organized as an option-preference matrix that represents player preferences over sequences of previous story events. This is called the Prefix Based Collaborative Filtering (PBCF) approach. This matrix is used to train a player model for players with different preferences. The trained model is then used by the game manager to choose among different options in each plot point.

The main disadvantage of this approach is that it is necessary to ask the player to explicitly rate the game manager options. In our approach, we don't need to ask anything to the actor of the observed behaviour. Preferences (as captured by the relations *desired*, *neutral* and *undesired*) are learned from the observed behaviour without any supervision.

In [Shen *et al* 2010], player preferences have an impact on the game dynamics. The player preferences are acquired through data mining techniques. The paper is not specific regarding the data mining techniques used to learn the player preferences thus we cannot compare it with ours.

[Baker, Saxe and Tenenbaum 2009][Riordan *et al* 2011][Tauber and Steyvers 2011] propose beliefs, goals and preferences (priorities) as fundamental concepts required for understanding the agent behaviour. In [Riordan *et al* 2011], preference is presented as a higher-level motive (e.g., avoiding zones with bad weather) that can be met by more concrete alternative goals. We interpret this notion of preference as a persistent higher-level goal as opposed to operational lower level ones. In our previous work [Botelho et al 2019], agents have beliefs (those provided by their sensors) and are capable of learning that they have goals. The same algorithms we have proposed for an agent to discover its goals from observation could be used to infer the user's goals.

[Baker, Saxe and Tenenbaum 2009][Riordan *et al* 2011][Tauber and Steyvers 2011] use Bayesian inverse planning for determining the beliefs and goals that may explain observed behaviours. This approach relies on the assumption that the observed behaviour is the result of a rational choice process in which the observed actor rationally chooses the best action to achieve their goals. Basically, the approach consists of inverting a planning problem in order to determine the probability that the observed actor has certain goals and beliefs given that a certain action was observed. This approach requires previous knowledge of the domain (probabilistic dynamics by which the observed actor moves through the environment and about the actor's set of possible goals). Contrarily to the described approach, ours does not require any previous knowledge of the domain.

[Baker, Saxe and Tenenbaum 2009] presents and tests three different inverse planning models and a fourth heuristic model in different scenario configurations. The tests reveal that one model (in which the goals of the observed actor may change) is better for some configurations while another one (in which the observed actor may have sub goals) is better for other cases. This means that, in spite the fact that these two models accurately predict the way people make inferences about explanations of the observed behaviour, the approach does not model the way people switch from one way of making inferences to the other.

[Shen et al 2010] proposes the representation of two additional concepts: Goal / Sub goal relations, and goal transitions as a result of action sequences. As discussed in section 2.1.10, about the RST relation *elaboration*, goal/sub goal (or intention / sub intention) relations arise of the hierarchic structure of complex behaviour. Goal transitions resulting from action sequences also reflect the hierarchic structure of complex behaviour. The relations *achieved* and *contributed*, originally proposed in our previous paper [Botelho et al 2019], describe such hierarchic structure.

The first two of these, *Achieved* and *Contributed*, are applicable to specific behaviours. *Achieved* represents the observed fact that a certain action achieved one or more of the agent goals, or one or more preconditions of actions that were subsequently executed in the same behaviour. *Contributed* represents the observed fact that a certain proposition contributed to the execution of an action because it is one of its preconditions. Although our previous work [Botelho et al 2019] explicitly addressed the means for an agent to describe its own behaviour, the proposed axiom schemata and algorithms could as well be used to describe the behaviour of some other agent, including the user.

Player emotions have also been used by drama managers to better fit their choices to the player [Shen *et al* 2010]. The player emotions are acquired through the analysis of the images provided by a camera in combination with the OCC model [Ortony, Clore and Collins 1988], which is a knowledge-based model of emotion elicitation.

We can imagine that emotion recognition through image analysis, used in [Shen *et al* 2010], does not require any domain dependent knowledge (it may rely on facial expressions that, according to Paul Ekman [Ekman 1992], are transversal across cultures and ethnic factors).

Our work has only addressed an epistemic emotion we call perplexity. Other emotions were not considered. The literature on epistemic emotions such as perplexity is reviewed latter.

*Incompatibility between propositions and/or actions*. Using our proposals, an agent can learn, just by observing behaviour, that two propositions and/or actions are mutually exclusive (relation *incompatible*). Incompatibility between propositions or between actions is an important information for planning algorithms; incompatibility between actions (events) has been used in automatic story generation systems.

For instance, the well-known *Graphplan* algorithm [Blum and Furst 1997] creates a graph consisting of a sequence of interleaved proposition layers and action layers. Both types of layers contain mutual exclusion relations, either between propositions or between actions. The algorithm uses a possibly incomplete set of rules to discover such pairs of mutually exclusive entities. Although the algorithms we propose for an agent to learn incompatibility between propositions or between actions are not adequate to be used in a planning algorithm, this reveals the importance of the learned concept of incompatibility.

[Sina, Rosenfeld and Kraus 2014] describes an approach for identifying events, in a story, that are inconsistent with available data. The algorithms proposed in [McIntyre and Lapata 2010] ensure that the events in a generated story satisfy all applicable temporal and mutual exclusion constraints. In [Li *et al* 2014] [Li *et al* 2013], the events in the learned story plot graphs are also subject to temporal ordering constraints and mutually exclusion relations. The present article considered the incompatibility between actions (events). In previous work [Botelho et al 2019] we describe the means for an agent to learn, by observation, that while it is trying to achieve its goal, it must reach a world state containing a certain proposition (e.g., $P_1$) before it reaches a world state containing another proposition (e.g., $P_2$). These temporal precedence constraints is represented by the relation *MustPrecede* (e.g., MustPrecede($P_1$, $P_2$)).

We also propose a stronger concept of incompatibility, captured by the relation *Prevents/2*, meaning that a proposition, if true in a given state, prevents the future occurrence of another proposition or action, or that an action, executed in a given state, prevents the future occurrence of another action or proposition.

[Chen at al 2009] propose a similar concept of incompatibility (londex, long-distance exclusion), which they apply to different versions of Satplan algorithm [Kautz and Selman 1992]. The relationship *londex* between propositions or between actions captures the idea that two propositions or two actions belonging to different plan steps are mutually exclusive. Although the relation *prevents*, we have proposed, captures the long-distance exclusion relation used in [Chen at al 2009], the algorithm we have used for an agent to learn it by observation does not apply to planning. However, the use of *londex* relations is evidence of the importance of our relation *prevents*.

*Perplexity*. In this paper, we provided the means for an agent to learn that a certain action of an observed behaviour causes perplexity because it leads the actor from a certain state of affairs to another seemingly worse state, according to some ideality principle, that is, a principle that the actor of the observed behaviour ought to comply with (in order to achieve its goal).

According to literature on philosophy [de Cruz 2021], social sciences [Deckert and Koenig 2017], psychology [Vogl et al 2019] and cognitive science [McPhetres 2019], emotions as perplexity, surprise and awe are epistemic emotions in the sense that they motivate the person feeling them to seek the new knowledge required to understand the cause of the emotion.

We couldn't find research on using perplexity either as an aspect of behaviour description or to improve a computer system's behaviour. However, given the above literature, a perplexed agent would be expected to initiate behaviour that would allow it to gain the knowledge required to cope with the causes of its perplexity, for instance, by engaging in a more exploratory mode of behaviour.

[Reizinger and Szemenyei 2020] and [Pathak et al 2017] describe reinforcement learning approaches where the agent involves itself in exploratory behaviour.

Putting all the above together, we think that when an agent becomes puzzled or perplexed with its own behaviour, it could use reinforcement learning approaches similar to those described in [Reizinger and Szemenyei 2020], through which it would be capable of overcoming the behavioural limitations that have caused it to be perplexed with its own behaviour. This approach does not apply when the observed behaviour is not the observing agent's own behaviour because reinforcement learning is used by the agent to control its own actions, not those of a different actor.

*Relation definition*. In this paper, we provide an algorithm that may be used by an agent to learn proposition definitions from the observed behaviour (relation *defining*). Defining(Prop, PropSet) means the proposition *Prop* is defined as the conjunction of all propositions in the set *PropSet*. For instance, in the Blocks World, if the agent's sensors are sensible to the propositions On(block1, block2), Clear(block) and Stack(stack), the algorithm learns that Stack([A, B, C]) is defined as the conjunction On(A, B) ∧ On(B, C) ∧ Clear(C).

Although we haven't found any research on using proposition definitions to describe the environment of an artificial agent or to improve its behaviour, it would be possible to use inductive logic programming [Muggleton and Raedt 1994] or decision tree learning [Quinlan 1986] to learn an implication as the following:

On(A, B) ∧ On(B, C) ∧ Clear(A) ⇒ Stack([A, B, C])[4]

---

[4] Decision trees do not work with variables but A, B and C are constants denoting specific blocks.

Our proposal has some advantages when compared with inductive logic programming or decision tree learning. First, while inductive logic programming and decision tree learning are supervised learning approaches, ours is totally unsupervised. Second, in our approach, the agent does not ask itself what is the definition of Stack([A, B, C]). The algorithm we have proposed does not need to be told what it is looking for. It just discovers that, within all its observations, Stack([A, B, C]) is defined as the conjunction On(A, B) ∧ On(B, C) ∧ Clear(A) by looking at the co-occurrence of all the involved literals and by realizing that there are states of the world that preclude alternative possibilities such as On(A, B) being defined as the conjunction Stack([A, B, C]) ∧ On(B, C) ∧ Clear(A).

*Use of background knowledge*. In this paper, we have also provided an inference rule to be used by the agent to determine the prior background knowledge it has used to infer a certain relationship, in addition to observations it made (i.e., a set of propositions of its own). For instance, to infer Incompatible(On(A, B), Clear(B)) from the hypothesis On(A, B), the agent needs the prior background knowledge ∀x∀y On(x, y) ⇒ ¬Clear(y) (as well as the definition of incompatibility). This is represented by the following relationship:

Background({∀x∀y On(x, y) ⇒ ¬Clear(y)}, Incompatible(On(A, B), Clear(B))).

We haven't found any research on using the relation *background* to describe the environment of an artificial agent or to improve its behaviour. However, considering the inference rule used to infer it (section 3.1), we could use an automatic theorem prover. That's exactly what we have done by using SWI-Prolog[5] to make the necessary inferences. The comparison of SWI-Prolog with alternative theorem provers is out of the scope of this article.

# 5  Conclusions and future research

In [Botelho et al 2019], we have proposed a set of relations that the agent may use to describe observed behaviour, its actor and environment. We put forth a set of deduction axiom schemata and domain independent algorithms the agent uses to acquire the referred descriptions, just by observation. Here, we have proposed a considerable extension of the initially proposed set of relations, inspired in the Rhetorical Structure Theory (RST). We have also presented deduction axiom schemata and corresponding domain independent algorithms for the newly proposed relations. Finally, we have presented a proof of concept using an implemented agent in a concrete demonstration scenario where the agent was able to describe the observed behaviours, in terms of all the relations proposed in this article.

In general, the capability to describe behaviour may be seen as contributing to the currently most demanded and sought for explainable AI. And, the analysis of related research shows that several of the concepts we propose, which can be learned and used by the agent to describe observed behaviour, its actor and environment were also used in other problems. This shows that our particular choices are important in specific areas.

Of the newly proposed relations, those involving more work and consideration are the relations of the family *Although*, inspired in the RST relation *Concession*. This family of relations empowers the agent with the capability of becoming perplexed with the observed behaviour (either its own or that of some other agent, including the user) and, sometimes finding an explanation for the surprising actions. The accurate definition of the

---

[5] SWI-Prolog: https://www.swi-prolog.org/

*Although* family of relations relied on several deontic concepts, whose definition and corresponding algorithms have been presented in this article.

Our proposal is probably the first to address perplexity as a way to describe observed behaviour, especially the means we have proposed for an agent to identify potentially perplexing actions, although some may be justified (using the relations *Although/4* and *Although/5*).

Even though the relations of the family *Although* already constitute an innovative and sophisticated contribution of our work, the word "although" (and other linguistic constructs with similar pragmatics) may be used in discourse in very different ways but causing perplexity. In future work, we will consider other ways the construct may be used.

From the analysis of the Rhetorical Structure Theory (RST), we propose the following useful set of relations, which empower the agent with a richer understanding of the observed behaviour, its actor and environment:

| NextState | Effects (Positive and negative effects) |
|---|---|
| Alternative | Precondition |
| Achieved | Incompatible |
| Contributed | Prevents |
| Although | Defining |
| Evaluation | Background |
| Evidence | |

Most of these RST-based relations are covered in previous work. However, three RST-based relations are yet not covered, namely *Alternative*, *Evaluation* and *Evidence*. *Alternative* may be used by the agent to represent alternative courses of action that solve the same problem.

*Evaluation* may be used by the agent to evaluate states of affairs in general and alternative courses of action in particular, and to choose one among them. Although we haven't actually tackled the relation *Evaluation*, we could have defined it from the relation *Desired*, *Neutral* and *Undesired*, which apply to propositions and actions. Defined this way, *Evaluation* would represent a three-valued scale of qualitative preferences.

*Evidence* could be used by the agent to represent evidence that a certain proposition is true. The evidence corresponds to a set of propositions: Evidence(PropSet, Prop), which may informally be read as "The propositions in the set *PropSet* are evidence that the proposition *Prop* is true". This would allow the agent to reason about certain propositions, of which it may not be certain, in terms of the evidence it has observed for them. Likewise, it could also be used by an agent to reason about uncertainty with respect to the user.

The analysis of related literature suggests the use of the concept of optional behaviour. A sequence of actions is optional if it can but does not have to be incorporated in the observed behaviour. Although apparently useful for describing specific instances of behaviours, the concept of an optional sequence of actions also constitutes a problem because, in each state, there may be infinite many sequences of actions that lead the agent to the same state. To avoid this infinity problem, it would only make sense to identify optional sequences of actions among those that have actually been observed. Besides, many if not all optional sequences of actions could be better interpreted as useless (suboptimal) rather than optional because, in fact, they would not be required to achieve the behaviour's goals.

Our future work will address the accurate definition of the relations *alternative*, *evaluation* and *evidence* and the axiom schemata and the algorithms that may be used by the agent to acquire descriptions based on them. To learn that there are alternative courses of action, the agent needs to observe different action sequences to solve the same problem (same initial state and same final state).

The relation *evaluation* may readily be defined from the relations *desired*, *neutral* and *undesired* and extended to action sequences. However, it is possible that a qualitative scale of preferences with more than three values would be necessary in some applications. One possibility that immediately pops into mind is to make use of the ideality principles we have defined for the relations of the family *although* together with the importance ordering we have used among them.

As for the relation *evidence*, we could use a statistical adaptation of the algorithm used to learn proposition definitions (i.e., relation *defining*). Rather than trying to find the propositions that always co-occur with another proposition, the algorithm would discover the propositions that often (but not always) co-occur with the proposition they support.

We have not proposed the concept of optional event, state, or course of action, but, from our analysis of the RST theory and the literature on automatic story-plot generation, we recognized its importance. We will also consider the possibility and usefulness of a restricted form of the concept of optional behaviour.

The reviewed literature revealed the importance of identifying emotions from the observed behaviours. The only emotion we have addressed was perplexity but others are certainly more important, especially in user modelling tasks. Another direction of future research is to identify other emotions from the observed behaviour.

We will also work on ways for the agent to use the acquired descriptions to improve its behaviour. For instance, when the agent has no justification for a perplexing action, it should engage in some form of exploratory behaviour such as using curiosity-driven reinforcement learning algorithms as those proposed in [Reizinger and Szemenyei 2019] and [Pathak 2017].

Although the algorithms we have proposed have the advantage of being unsupervised and not requiring any previous knowledge about the specific application domain, they rely on the observation of the behaviours that achieve the same goal, starting from all possible configurations of the initial state. This is a major drawback because often the agent would need to use its acquired descriptions way before it has the chance to observe the behaviours that achieve a certain goal for all possible initial states. It is even likely that, in many problems, there are a huge number of initial states making it unpractical to explore all of them. The way around this demands algorithms that would incrementally learn the mentioned descriptions, even if only approximate ones. We believe that after some iterations, the incrementally learned descriptions may be good enough or even the final descriptions.

Another major drawback of our work consists of assuming that the final states of all observed behaviours are successful states in which the pursued goals are satisfied. We recognize that this is not a reasonable assumption for many problem classes.

Defining incremental algorithms and dropping the hypothesis that the final states of all observed behaviour are successful will be our most important concerns for future research. This will make our approach more robust and more widely usable.

Incremental learning will enable the agent to use its descriptions in problems in which our approach is not yet competitive. This might be the case if the agent tries to use the learned relations *incompatible* and *prevents* in planning algorithms.

# 6   References

[Baker, Saxe and Tenenbaum 2009] Baker, C.L.; Saxe, R.; Tenenbaum, J.B. 2009. Action understanding as inverse planning. *Cognition*, 113:329-349. doi:10.1016/j.cognition.2009.07.005

[Barocas et al 2017] Barocas, S.; Crawford, K.; Shapiro, A.; and Wallach, H. 2017. The problem with bias: from allocative to representational harms in machine learning Special Interest Group for Computing, Information and Society (SIGCIS)

[Blum and Furst 1997] Blum, A.; and Furst, M. 1997. Fast planning through planning graph analysis. Artificial intelligence. 90(1-2):281-300. DOI: 10.1016/S0004-3702(96)00047-1

[Botelho *et al* 2019] Botelho, L.M.; Lopes, R.; Nunes, L.; Ribeiro, R. 2019. Software agents with concerns of their own. Submitted to the *Cognitive Science Journal*

[Buchsbaum *et al* 2015] Buchsbaum, D.; Griffiths, T.L.; Plunkett, D.; Gopnik, A.; Baldwin, D. 2015. Inferring action structure and causal relationships. *Cognitive Psychology*, 76:30–77

[Cerone 2019] Cerone, A. 2019. Behaviour and Reasoning Description Language (BRDL). In: Camara J.; and Steffen M. (eds) Software Engineering and Formal Methods (SEFM 2019). Lecture Notes in Computer Science: Vol 12226. Springer, Cham. pp 137-153. DOI: 10.1007/978-3-030-57506-9_11

[Chen at al 2009] Chen, Y.; Huang, R; Xing, Z.; and Zhang, W. 2009. Long-Distance Mutual Exclusion for Planning. Artificial Intelligence 173(2):365–391. DOI: 10.1016/j.artint.2008.11.004

[de Cruz 2021] de Cruz, H. 2021 Perplexity and Philosophical Progress. Midwest Studies in Philosophy 45:209-221. DOI: 10.5840/msp20219166

[Deckert and Koenig 2017] Deckert, J.C.; and Koenig, T.L. 2017. Social work perplexity: Dissonance, uncertainty, and growth in Kazakhstan. Qualitative Social Work 18(2):163-178 DOI: 10.1177/1473325017710086

[Ekman 1992] Ekman, P. 1992. An argument for basic emotions. *Cognition and Emotion*, 6(3-4):169-200

[Freed and Remington 2000] Freed, M.; and Remington, R. 2000 GOMS, GOMS+, and PDL. AAAI Technical Report FS-00-03

[Gabbay *et al* 2013] Gabbay, D.; Horty, J.; Parent, X.; van der Meyden, R.; van der Torre, L. (Editors) 2013. Handbook of Deontic Logic and Normative Systems. Volume 1. College Publications. ISBN 978-1-84890-132-2

[Ghallab, Nau and Traverso 2004] Ghallab, M.; Nau, D.; and Traverso, P. 2004. Automated Planning: Theory and Practice. The Morgan Kaufmann Series in Artificial Intelligence. ISBN 978-1-55860-856-6

[Harrison *et al* 2014] Harrison, B.; Ware, S.G.; Fendt, M.W.; Roberts, D.L. 2014. A survey and analysis of techniques for player behavior prediction in massively multiplayer online games. *IEEE Transactions on Emerging Topics in Computing Special Issue on MMO Technologies*. DOI: 10.1109/TETC.2014.2360463

[Hendrikx *et al* 2013] Hendrikx, M.; Meijer, S.; van der Velden, J.; Iosup, A. 2013. Procedural content generation for games: a survey. *ACM Transactions on Multimedia Computing, Communications, and Applications*, 9(1):1-22. DOI: 10.1145/2422956.2422957

[Hilpinen 1981] Hilpinen, R. (Editor). 1981. New Studies in Deontic Logic. D.Reidel. Dordrecht

[Kautz and Selman 1992] Kautz, H.A.; and Selman, B. 1992. Planning as satisfiability. In Proceedings of the Tenth European Conference on Artificial Intelligence (ECAI 1992):359-363

[Kelsen, Pulvermueller and Glodt 2007] Kelsen, P.; Pulvermueller, E.; and Glodt, C. 2007. A Declarative Executable Language based on OCL for Specifying the Behavior of Platform-Independent Models. Proceedings of 2007 Workshop Ocl4All: Modelling Systems with OCL (Ocl4All 2007).

[Köchling and Wehner 2020] Köchling, A.; and Wehner, M.C. 2020. Discriminated by an algorithm: a systematic review of discrimination and fairness by algorithmic decision-making in the context of HR recruitment and HR development. Business Research volume 13, pp795–848. DOI: 10.1007/s40685-020-00134-w

[Kosseim and Lapalme 2000] Kosseim, L.; Lapalme, G. 2000. Choosing Rhetorical Structures to Plan Instructional Texts. *Computational Intelligence*, 16(3):408-455

[Li *et al* 2013] Li, B.; Lee-Urban, S.; Johnston, G.; Riedl, M. 2013. Story generation with crowdsourced plot graphs. In *Proceedings of the National Conference on Artificial Intelligence* (AAAI 2013)

[Li *et al* 2014] Li, B.; Thakkar, M.; Wang, Y.; Riedl, M.O. 2014. Storytelling with Adjustable Narrator Styles and Sentiments. In *Proceedings of the 2014 International Conference on Interactive Digital Storytelling*. Singapore

[Mann 1984] Mann, W.C. 1984. Discourse Structures for Text Generation. In *Proceeding of the 10th International Conference on Computational Linguistics* (ACL 1984). p367-375. DOI: 10.3115/980431.980567

[Mann and Thompson 1987] Mann, W.C.; Thompson, S.A. 1987. Rhetorical Structure Theory: A Framework for the Analysis of Texts. *IPRA (International Pragmatics Association) Papers in Pragmatics* 1:1-21

[Mann and Thompson 1988] Mann, W.C.; Thompson, S.A. 1988. Rhetorical Structure Theory: Toward a functional theory of text organization. *Text*, 8(3):243-281

[Marcu 2000] Marcu, D. 2000. The Rhetorical Parsing of Unrestricted Texts: A Surface-based Approach. *Computational Linguistics*, 26(3):395-448

[McIntyre and Lapata 2010] McIntyre, N.; Lapata, M. 2010. Plot induction and evolutionary search for story generation. In *Proceedings of the 48th Annual Meeting of the Association for Computational Linguistics*. p1562-1572

[McPhetres 2019] McPhetres, J. 2019. Oh, the things you don't know: awe promotes awareness of knowledge gaps and science interest. *Cognition and Emotion* 33(8):1599-1615. DOI: 10.1080/02699931.2019.1585331

[Miller 2019] Miller, T. 2019. Explanation in artificial intelligence: Insights from the social sciences. Artificial Intelligence. Volume 267 pp. 1-38. DOI: 10.1016/j.artint.2018.07.007

[Muggleton and Raedt 1994] Muggleton, S.; and Raedt, L. 1994. Inductive Logic Programming: Theory and methods. The Journal of Logic Programming, Volumes 19–20, Supplement 1, Pages 629-679. DOI: 10.1016/0743-1066(94)90035-3

[Ortony, Clore and Collins 1988] Ortony, A.; Clore, G.L.; Collins, A. 1988. The Cognitive Structure of Emotions. Cambridge University Press. Cambridge, UK.

[Pathak et al 2017] Pathak; Agrawal; Efros; Darrell 2017 Curiosity-driven Exploration by Self-supervised Prediction. In Proceedings of the 2017 International Conference on Machine Learning (ICML 2017), Sydney, Australia. Volume 70. pp2778–2787 [Piller, Vincze and Kovács 2015] Piller, I.; Vincze, D.; and Kovács, S. 2015. Declarative Language for Behaviour Description. In Sinčák P., Hartono P., Virčíková M., Vaščák J., Jakša R. (eds.) Emergent Trends in Robotics and Intelligent Systems. Advances in Intelligent Systems and Computing 316. Springer. pp 103-112. DOI: 10.1007/978-3-319-10783-7_11

[Pinhanez et al 2021] Pinhanez, C.S.; Cavalin, P.; Ribeiro, V.; Appel, A.P.; Candello, H; Nogima, J.; Pichiliani, M; Guerra, M; de Bayser, M.G.; Malfatti, G.; and Ferreira, H. 2021. Using Meta-Knowledge Mined from Identifiers to Improve Intent Recognition in Conversational Systems. In Proceedings of the 59th Annual Meeting of the Association for Computational Linguistics and the 11th International Joint Conference on Natural Language Processing, pages 7014–7027. August 1–6, 2021. Association for Computational Linguistics

[Quinlan 1986] Quinlan, J.R. 1986. Induction of Decision Trees. *Machine Learning* 1:81-106. DOI: 10.1007/BF00116251

[Ramirez and Bulitko 2014] Ramirez, A.; Bulitko, V. 2014. Automated Planning and Player Modelling for Interactive Storytelling. *IEEE Transactions on Computational Intelligence and AI in Games*. DOI: 10.1109/TCIAIG.2014.2346690

[Reizinger and Szemenyei 2020] Reizinger, P.; and Szemenyei, M. 2019. Attention-based Curiosity-driven Exploration in Deep Reinforcement Learning. In Proceedings of 2020 IEEE International Conference on Acoustics, Speech and Signal Processing (ICASSP 2020). DOI: 10.1109/ICASSP40776.2020.9054546

[Riedl and Bulitko 2012] Riedl, M.O.; Bulitko, V. 2012. Interactive narrative: An intelligent systems approach. *AI Magazine*, 34(1):67-77. DOI: http://dx.doi.org/10.1609/aimag.v34i1.2449

[Riordan *et al* 2011] Riordan, B.; Bruni, S.; Schurr, N.; Freeman, J.; Ganberg, G; Cooke, N.J.; Rima, N. 2011. Inferring user intent with Bayesian inverse planning: Making sense of multi-UAS mission management. In *Proceedings of the 20th Behavior Representation in Modeling and Simulation Conference* (BRIMS). Sundance, UT

[Samek et al 2019] Samek, W.; Montavon, G.; Vedaldi, A.; Hansen, L.K.; and Müller, K-R. (Editors) 2019 Explainable AI: Interpreting, Explaining and Visualizing Deep Learning. Lecture Notes in Computer Science book series (LNCS, volume 11700). Lecture Notes in Artificial Intelligence book sub series (LNAI, volume 11700). Springer, Cham. Softcover ISBN: 978-3-030-28953-9. eBook ISBN: 978-3-030-28954-6. DOI: 10.1007/978-3-030-28954-6

[Shen *et al* 2010] Shen, Z.; Miao, C.; Zhang, L.; Yu, H.; Chavez, M.J. 2010. An emotion aware agent platform for interactive storytelling and gaming. In *Proceedings of the International Academic Conference on the Future of Game Design and Technology* (Futureplay 2010). p257-258. ACM. New York, NY, USA. DOI: 10.1145/1920778.1920823

[Sina, Rosenfeld and Kraus 2014] Sina, S.; Rosenfeld, A.; Kraus, S. 2014. Generating content for scenario-based serious games using crowdsourcing. In *Proceedings of the 28th National Conference on Artificial Intelligence (AAAI 2014)*

[Suresh and Guttag 2021] Suresh, H.; and Guttag, J.V. 2021. A Framework for Understanding Sources of Harm throughout the Machine Learning Life Cycle. In the Proceedings of the 2021 ACM Conference on Equity and Access in Algorithms, Mechanisms, and Optimization (EAAMO 2021). Article 17. DOI: 10.1145/3465416.3483305

[Tauber and Steyvers 2011] Tauber, S.; Steyvers, M. 2011. Using inverse planning and theory of mind for social goal inference. In *Proceedings of the 33rd Annual Conference of the Cognitive Science Society*. p2480-2485 [Thue, Bulitko and Spetch 2008] Thue, D.; Bulitko, V.; and Spetch, M. 2008. PaSSAGE: A demonstration of player modelling in interactive storytelling. In *Procedings of the Annual AAAI Conference on Artificial Intelligence and Interactive Digital Entertainment* (AIIDE 2008). p227-228

[Uzêda, Pardo and Nunes 2010] Uzêda, V.R.; Pardo, T.A.S.; Nunes, M.G.V. 2010. A comprehensive comparative evaluation of RST-based summarization methods. *ACM Transactions on Speech and Language Processing*, 6(4):1-20

[van der Waa et al 2021] van der Waa, J.; Nieuwburg, E.; Cremers, A.; and Neerincx, M. 2021. Evaluating XAI: A comparison of rule-based and example-based explanations. Artificial Intelligence. Volume 291 103404. DOI: 10.1016/j.artint.2020.103404

[Vogl et al 2019] Vogl, E.; Pekrun, R.; Murayama, K.; Loderer, K.; and Schubert, S. 2019. Surprise, Curiosity, and Confusion Promote Knowledge Exploration: Evidence for Robust Effects of Epistemic Emotions. Frontiers in Psychology 10, article 2474. DOI: 10.3389/fpsyg.2019.02474

[Wan, Wu and Wu 2009] Wan, L.; Wu, G.; and Wu, H. 2009. BDL – Behaviour Description Language. In Proceedings of the the International Conference on Software, Technology and Engineering. Iscte-IUL. pp. 37-41. DOI: 10.1142/9789814289986_0008

[Yu and Riedl 2012] Yu, H.; Riedl, M. O. 2012. A sequential recommendation approach for interactive personalized story generation. In *Proceedings of the International Conference on Autonomous Agents and MultiAgent Systems* (AAMAS 2012). p71-78

[Yu and Riedl 2013] Yu, H.; Riedl, M. O. 2013. Toward personalized guidance in interactive narratives. In *Proceedings of the International Conference of the Foundation of Digital Games* (FDG 2013)